\RequirePackage{fix-cm}
\documentclass[twocolumn]{svjour3}          
\usepackage{natbib}
\smartqed

\usepackage{times}
\usepackage{epsfig}
\usepackage{array}
\usepackage{graphicx}
\usepackage{amsmath}
\usepackage{amssymb}
\usepackage[colorlinks, citecolor=blue]{hyperref} 
\usepackage{textcomp}
\usepackage{bbm}
\usepackage{balance}
\usepackage[T1]{fontenc}    
\usepackage{microtype}  
\usepackage{ulem}
\usepackage{bm}
\usepackage{caption}
\usepackage{subfigure}
\usepackage{booktabs}
\usepackage{multirow}
\usepackage{makecell}
\usepackage{color}

\newcommand{\squishlist}{
	\begin{list}{$\bullet$}
		{ \setlength{\itemsep}{0pt}
			\setlength{\parsep}{2pt}
			\setlength{\topsep}{2pt}
			\setlength{\partopsep}{0pt}
			\setlength{\leftmargin}{1em}
			\setlength{\labelwidth}{1em}
			\setlength{\labelsep}{0.5em} } }
	\newcommand{\squishend}{
\end{list} }

\usepackage{colortbl}
\usepackage{tabularx}
\usepackage{xspace}

\usepackage[linesnumbered,ruled,vlined]{algorithm2e}
\usepackage{algpseudocode}

\makeatletter
\DeclareRobustCommand\onedot{\futurelet\@let@token\@onedot}
\def\@onedot{\ifx\@let@token.\else.\null\fi\xspace}
\def\ourmethod{{\textsc{Degree}}\xspace}
\def\eg{\emph{e.g}\onedot} 
\def\ie{\emph{i.e}\onedot}

\makeatother

\begin{document}

\title{A2B: Anchor to Barycentric Coordinate for Robust Correspondence\thanks{\textit{Corresponding author: Zhiguo Cao. 
				W. Zhao and H. Lu contributed equally.
			}			
		}
	}

	\author{Weiyue Zhao$^{1}$ \and  Hao Lu$^{1}$ \and Zhiguo Cao$^{1}$ \and Xin Li$^{2}$
	}

	\institute{
            Weiyue Zhao$^{1}$ \at
		\email{zhaoweiyue@hust.edu.cn}         		
		\and
		Hao Lu$^{1}$ \at
		\email{hlu@hust.edu.cn}
		\and
		Zhiguo Cao$^{1}$ \at
		\email{zgcao@hust.edu.cn}
		\and
		Xin Li$^{2}$ \at
		\email{xin.li@mail.wvu.edu}
		\\	
		\and $^{1}$	the Key Laboratory of Image Processing and Intelligent Control, Ministry of Education; School of Artificial Intelligence and Automation, Huazhong University of Science and Technology, Wuhan, 430074, China \\
		\\
		$^{2}$ Lane Department of Computer Science and Electrical Engineering, West Virginia University, Morgantown 15461, Commonwealth of Virginia, America\\
	}

	\date{Received: date / Accepted: date}

	\maketitle


\begin{abstract}{
There is a long-standing problem of repeated patterns in correspondence problems, where mismatches frequently occur because of inherent ambiguity. The unique position information associated with repeated patterns makes coordinate representations a useful supplement to appearance representations for improving feature correspondences. However, the issue of appropriate coordinate representation has remained unresolved. In this study, we demonstrate that geometric-invariant coordinate representations, such as barycentric coordinates, can significantly reduce mismatches between features. The first step is to establish a theoretical foundation for geometrically invariant coordinates. We present a seed matching and filtering network (SMFNet) that combines feature matching and consistency filtering with a coarse-to-fine matching strategy in order to acquire reliable sparse correspondences. We then introduce \ourmethod, a novel anchor-to-barycentric (A2B) coordinate encoding approach, which generates multiple affine-invariant correspondence coordinates from paired images. \ourmethod can be used as a plug-in with standard descriptors, feature matchers, and consistency filters to improve the matching quality. Extensive experiments in synthesized indoor and outdoor datasets demonstrate that \ourmethod alleviates the problem of repeated patterns and helps achieve state-of-the-art performance. Furthermore, \ourmethod also reports competitive performance in the third Image Matching Challenge at CVPR 2021. This approach offers a new perspective to alleviate the problem of repeated patterns and emphasizes the importance of choosing coordinate representations for feature correspondences.
}


\end{abstract}

\maketitle

\section{Introduction}

\begin{figure}[!t]
\centering
   \subfigure[\scshape{SuperPoint} + \scshape{SuperGlue}]
   {\includegraphics[width=0.23\textwidth]{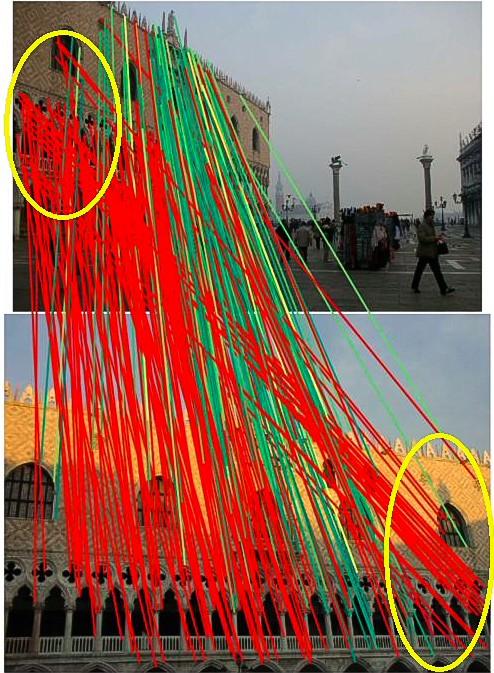}} 
    \subfigure[\scshape{SuperPoint} + \ourmethod + \scshape{SuperGlue}]{
    \includegraphics[width=0.23\textwidth]{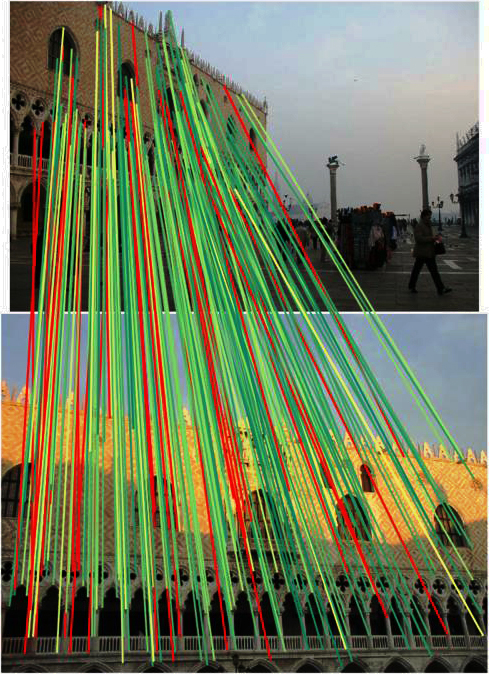}} 
\caption{Failure-matching cases due to repeated patterns. (a) are the results of the top entries in the CVPR 2020 Image Matching Challenge~\citep{jin2020image}. Red indicates matches above a 5-pixel error threshold, and those below are color-coded by their errors, from $0$ (green) to $5$ pixels (yellow). Note that there exist mismatches with significant spatial shifts (yellow circles). (b) our result with \ourmethod (note a significant improvement, as demonstrated by more green lines).}
\label{Fig:main}
\end{figure}

Correspondence is a fundamental problem in computer vision that serves many applications~\citep{lu2017deep,mur2015orb,schonberger2016structure,sun2008learning}. The search for consistent correspondences, \textit{a.k.a.} matches, includes two key steps: feature matching~\citep{baumberg2000reliable,5582232} and consistency filtering~\citep{bian2017gms,lowe2004distinctive,ma2019locality}. In classic pipelines, matching handcrafted descriptors such as SIFT~\citep{lowe2004distinctive} generates initial matches, which are then filtered by some operators, such as the mutual nearest-neighbor check and the ratio test~\citep{lowe2004distinctive}, to find reliable ones. Due to the limitation of local descriptors, classic methods often struggle to deal with challenging scenarios such as repeated patterns, low-texture regions, and symmetric structures. As shown in Fig.~\ref{Fig:main}(a), despite many correct correspondences, `SuperPoint + SuperGlue' still produce many outliers (as highlighted by yellow circles). 

To improve the robustness of feature correspondences, in this work, we propose two novel strategies: {\em reliable seed correspondence finding} and {\em geometric invariant coordinate encoding}. The first is to acquire reliable {\em seed} correspondences (initial guess) using the coarse-to-fine matching strategy \citep{sun2021loftr}. In contrast to existing practices that solve feature matching and consistency filtering separately, we propose to jointly optimize the two tasks in a single network called Seed Matching and Filtering Network (SMFNet). The key idea is to first generate seed correspondences using the Sinkhorn algorithm~\citep{cuturi2013sinkhorn} at the coarse level. We can then refine the matching results by estimating the compatibility based on self-attention. Iterative rejection of outliers is implemented by designing a new neighbor mining block (NM-block)~\citep{zhao2019nm} and an attentive residual block (AR-block)~\citep{sun2020acne}. 
The generated seed correspondences will be used to build multiple coordinate systems and serve our second goal: geometric invariant coordinate encoding.

Novel strategies have been proposed to integrate the coordinate information of keypoints to improve the performance of the matching ~\citep{rocco2018neighbourhood,sun2020acne,wang2019deep,yi2018learning}. 
Despite the benefit of coordinate encoding, the problem of correspondences still faces challenges in certain scenarios, even with high-quality descriptors such as SuperPoint~\citep{detone2018superpoint} (Fig.~\ref{Fig:main}(a)). Are the descriptors not representative? We are afraid that this is not the case. 
By considering another baseline `HardNet~\citep{mishchuk2017working}+NMNet~\citep{zhao2019nm}' that does not encode any coordinate/global information, we observe that 
many correspondences all make sense locally, but only 
few correspondences are correct from a global view. It follows that \textit{inappropriate coordinate encoding leads to unreliable feature correspondences}. Indeed, it is well known that Cartesian or polar coordinates are sensitive to scale, rotation, and affine transformations. It is therefore natural to ask: \textit{Can we improve feature correspondences with geometric-invariant coordinates?} 

The answer is yes. In this work, we 
study the blessings of the barycentric coordinate system~\citep{li2019lam} for the learned correspondence. 
In particular, we present \ourmethod, a novel coordinate encoding approach that implements Dynamic Encoding in the Geometric-invaRiant coordinatE systEm (\ourmethod), for feature correspondences. Following~\cite{li2019lam}, we also use the barycentric coordinate system~\citep{marschner2018fundamentals} due to its invariance property, but a key attribute of \ourmethod is that \ourmethod encodes coordinates conditioned on dynamic and sparse correspondences from paired images and based on a set of barycentric coordinate systems. Specifically, each system is constructed with a set of bases formed by $3$ basis points. The basis points are chosen from some reliable, sparse seed correspondences. 
In this way, descriptors/features can embed positional information with a correspondence-specific coordinate vector. Compared 
with using conventional coordinate representations, \ourmethod shows the potential to address the problem of repeated patterns (Fig.~\ref{Fig:main}(b)). 

To encode reliable coordinates, another issue is how to acquire reliable seed correspondences. Poor seed correspondences can affect coordinate encoding. In contrast to existing practices that solve feature matching or consistency filtering alone, we propose to jointly optimize the two tasks in a single network called Seed Matching and Filtering Network (SMFNet) to generate seed correspondences using a coarse-to-fine matching strategy (Fig.~\ref{fig:seed correspondences}). Furthermore, we proposed a preprocessing strategy for input descriptors to alleviate the interference of repeated patterns and ensure the efficiency and accuracy of SMFNet (see Sec.~\ref{sec:ablation}). It aims to acquire reliable seed correspondences (anchors) from non-repetitive regions, and subsequently, the obtained anchors can be used to encode coordinate representations to discriminate repetitive patterns between image pairs (Fig.~\ref{fig:seed correspondences}).

Extensive experiments 
on a synthesized toy dataset, an indoor dataset (SUN3D~\citep{xiao2013sun3d}), and two outdoor datasets (YFCC100M~\citep{thomee2016yfcc100m} and PhotoTourism~\citep{jin2020image}) show that i) \ourmethod effectively mitigates the problem of repeated patterns, that ii) \ourmethod helps achieve state-of-the-art matching performance, for example, it can work with the recent feature matcher SuperGlue~\citep{sarlin2020superglue} and the consistency filter OANet~\citep{zhang2019learning} to improve matching precision and recall (\ourmethod improves SuperGlue by at least $5.5\%$ in all metrics with HardNet), and that iii) \ourmethod is compatible with standard feature descriptors, such as RootSIFT~\citep{arandjelovic2012three}, HardNet~\citep{mishchuk2017working}, and SuperPoint~\citep{detone2018superpoint}. We also performed ablation studies to justify our design choices and validate the reliability of the seed correspondences generated. Furthermore, \ourmethod plays an important role in our CVPR 2021 image matching challenge entries, where we report competitive performance against other competitors on three public benchmarks. In particular, we 
rank the first on the GoogleUrban track with restricted keypoints.

Our main contributions include the following.
\begin{itemize}

\item We provide a novel perspective to alleviate the problem of repeated patterns in correspondence: resorting to non-repetitive regions to seek a few reliable seed correspondences (anchors) first, then constructing paired, geometric-invariant coordinate systems using the anchors, and finally using coordinate representations to discriminate repeated patterns;
\item We present \ourmethod: a novel anchor-to-barycentric coordinate encoding approach that i) finds reliable anchors with a light-weight and coarse-to-fine sparse correspondence generation network SMFNet, that ii) constructs multiple paired barycentric coordinate systems from the anchors, and that iii) encodes geometric-invariant barycentric coordinates for feature correspondence; \ourmethod is effective, easy-to-implement, and robust for real-world $3$D transformations;
\item We show that geometric-invariant coordinate representations point out a new direction from considering robust feature correspondences, \eg, \ourmethod can be used as a plug-in applicable to existing feature descriptors, the feature matcher, and the consistency filter to consistently improve the correspondence performance, particularly for learned correspondences.

\end{itemize}


\section{Related Work}
\subsection{Feature Descriptors}
Local descriptors can be categorized as handcrafted and learning-based. 
Here, we divide them into local-only descriptors and local-global descriptors. Local-only descriptors only encode local patterns and are typically extracted from image patches~\citep{mishchuk2017working,rosten2006machine,rublee2011orb}. However, when repeated patterns appear at the image level, they may not work well. Local-global descriptors encode both local and global signals. With the global cue, the descriptors can be discriminative even with similar local patterns. In particular, global cues can be pixel coordinates, context~\citep{luo2019contextdesc}, or semantic information. If feature extraction includes hierarchical image encoding~\citep{detone2018superpoint,liu2019gift}, feature grouping, and image pyramid~\citep{arandjelovic2012three,lowe2004distinctive}, such descriptors could also be considered local-global, because a larger receptive field leads to increased discrimination to spatially distant points. 
Despite various descriptors, they encode per-image features, but neglect information dependency between image pairs.

\subsection{Feature Correspondences}
In general, finding feature correspondences has two stages: feature matching and consistency filtering. Classic pipelines often build a coarse matching set according to measured feature distance and use criteria such as a fixed threshold, nearest neighbor (NN) and mutual NN. Consistency filtering methods can be parametric or nonparametric. The former includes \eg, RANSAC~\citep{fischler1981random}, PROSAC~\citep{chum2005matching}, LORANSAC~\citep{chum2003locally}), and the latter has, to name a few,
Grid-based motion statistics (GMS)~\citep{bian2017gms}, SGA~\citep{1544893}, vector field consensus (VFC)~\citep{ma2014robust}. These methods generally measure the Euclidean distance in the feature space or in the image coordinate system.

Recent work focuses on developing point-based matching strategies. Some explore the learnable rejection of outliers~\citep{ranftl2018deep,yi2018learning,zhang2019learning,zhao2020image}; Some attempt to find reliable matches \eg,~\cite{zhao2019nm} proposes a compatibility metric to find matches from the feature neighbors. The input of these methods requires the coordinates of the paired points. Moreover, the work of~\citep{sarlin2020superglue} proposes a learnable matching network SuperGlue with the Sinkhorn algorithm~\citep{cuturi2013sinkhorn,sinkhorn1967concerning}. Some recent methods~\citep{sgmnet,clustergnn} build on SuperGlue~\citep{sarlin2020superglue} to further improve computational efficiency and matching performance. In particular, SGMNet~\citep{sgmnet} establishes a small set of nodes to reduce the cost of attention. ClusterGNN~\citep{clustergnn} uses K-means to construct local subgraphs to save memory cost and computational overhead. It uses the position of the keypoint in the camera coordinate system. However, the Cartesian coordinate system is sensitive to geometric transforms. Neither stacked convolution nor fully connected layers can overcome these transformations.
In contrast to the Cartesian coordinate system, we propose using seed correspondences to build a set of barycentric coordinate systems with geometric invariance, which has shown to greatly improve the robustness of correspondences. 

\subsection{Coordinate Representations}
Coordinate representations in deep networks can be divided into the encoding of {\em absolute} and {\em relative} positions. The absolute position encoding was first proposed by~\cite{vaswani2017attention}. Since self-attention does not model position information explicitly, an absolute position representation is added to the input. 
Several choices are available for absolute position encoding, such as fixed encoding using multifrequency sine / cosine functions~\citep{sun2021loftr,vaswani2017attention} and learnable encoding with a network~\citep{jin2020image,sun2020acne,zhao2019nm}. In addition to the absolute position, recent work~\citep{DBLP,wu2021rethinking} also considers the relation between relative positions. For example, the latest representative work, SuperGlue~\citep{sarlin2020superglue}, proposes a keypoint encoder to mimic the `positional encoder used in natural language processing~\citep{gehring2017convolutional,vaswani2017attention} where two-dimensional keypoint coordinates are embedded in feature descriptors. However, almost all existing practices only consider the image coordinate system or the sequence position, which is sensitive to geometric transformation. 


\section{Learning to Seek Reliable Correspondences} \label{sec:seed choosen}
Encoding accurate correspondence coordinates requires accurate seed correspondences. To acquire reliable seed correspondences needed for constructing barycentric coordinate systems, we present a seed matching and filtering network (SMFNet) that jointly solves the problems of feature matching and consistency filtering.

\subsection{Seed Matching and Filtering Network (SMFNet)}
\begin{figure*}[!t]
\centering  
\subfigure[Network Architecture]{
\label{sub.1}
\includegraphics[width=\textwidth]{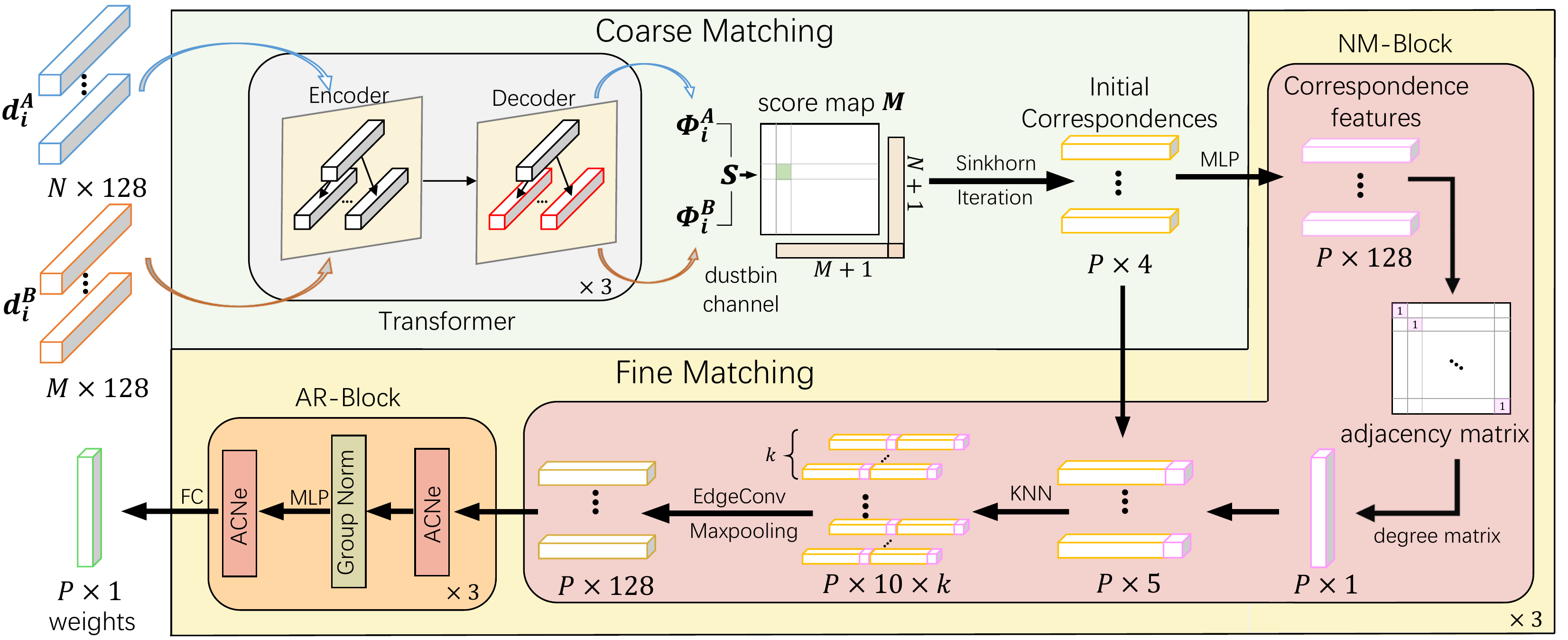}}
\quad
\subfigure[Key Modules]{
\label{sub.2}
\includegraphics[width=0.70\textwidth]{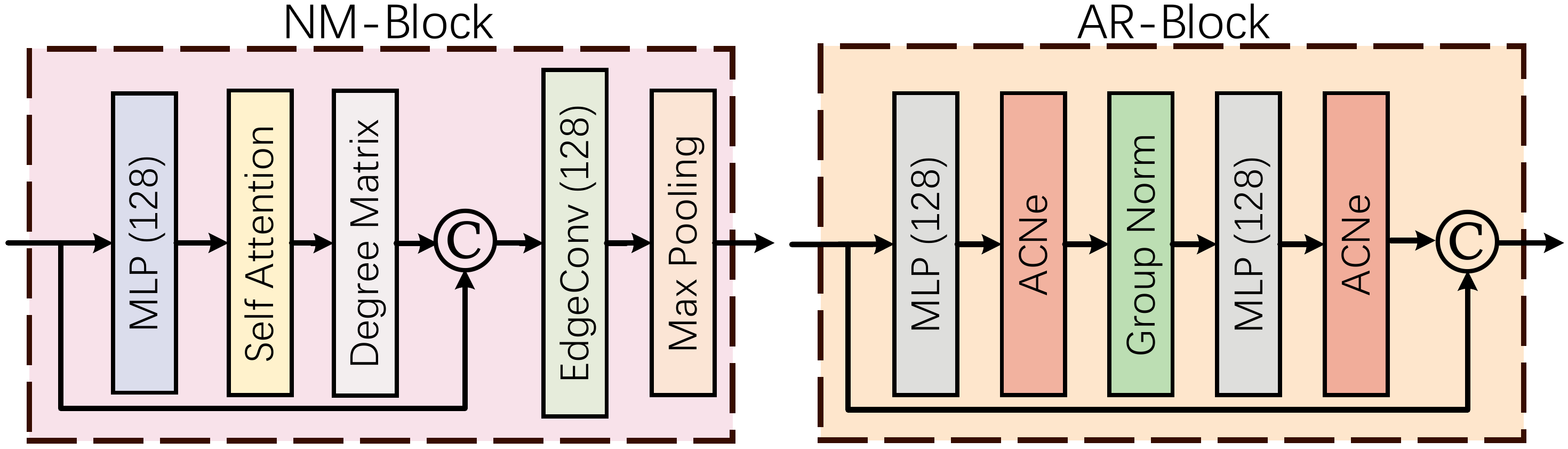}}
\caption{Architecture of SMFNet. (a) is the overall architecture. The input is two sets of descriptor vectors, and the SMFNet outputs reliable correspondences conditioned on their confidence values. $k$ is the hyperparameter of K-Nearest-Neighbor (KNN). (b) highlights two key modules: NM-Block and AR-Block.}
\label{Fig:network}
\end{figure*}

Since the geometric coordinate system is built from seed correspondences, the quality of these correspondences matters. One may consider using off-the-shelf feature matchers and/or consistency filters to generate the correspondences. However, existing methods~\citep{sun2021loftr,d2net2019CVPR} generally aim to find a moderate number of correspondences, but what we need are only a few but high-quality correspondences as basis points. To meet our specialized demand, we develop a light-weight network, termed the Seed Matching and Filtering Network (SMFNet), that mimics state-of-the-art feature matchers and consistency filters and that only generates sparse correspondences. In particular, unlike the existing work that solves each task individually, feature matching and consistency filtering are jointly optimized in SMFNet. Its architecture is shown in Fig.~\ref{Fig:network}(a). Inspired by the recent work of LoFTR \citep{sun2021loftr}, we designed a coarse-to-fine matching strategy to find seed correspondences such that coarse matching generates initial correspondences and fine matching further removes outliers.

\subsection{Coarse-level Matching} \label{subsec:coarse-level matchinag}
Inspired by~\citet{vaswani2017attention} and~\citet{wang2019deep}, we apply a transformer without positional encoding to encode the input local descriptor $\bm d_i^A$/$\bm d_i^B$ and to generate $\Phi_i^A$/$\Phi_i^B$, as
\begin{equation}
\begin{split}
    \Phi_i^{A(1)} &= \bm d_i^A + G_\text{self}(\bm d_i^A, \bm d_i^A)\,,\\
    \Phi_i^{B(1)} &= \bm d_i^B + G_\text{self}(\bm d_i^B, \bm d_i^B)\,, \\
    \Phi_i^{A(2)} &= \Phi_i^{A(1)} + G_\text{cross}(\Phi_i^{A(1)}, \Phi_i^{B(1)})\,, \\
    \Phi_i^{B(2)} &= \Phi_i^{B(1)} + G_\text{cross}(\Phi_i^{B(1)}, \Phi_i^{A(1)})\,,
\end{split}
\end{equation}
where $G_{self}$ and $G_{cross}$ represent self-attention and cross-attention with multihead $ H = 4$, respectively. As shown in Fig.~\ref{sub.1}, we adapt $3$ layers of alternated self- and cross-attention. Following~\cite{sarlin2020superglue}, we also apply the Sinkhorn algorithm~\citep{cuturi2013sinkhorn} (for $50$ iterations) to the score map $S=\Phi_i^A({\Phi_i^B})^T$ embedded with a dustbin channel, by adding another row and column filled with a single learnable parameter, to produce an assignment matrix ${\bm M} \in \mathbb{R}^{(N_A + 1) \times (N_B + 1)}$, as
\begin{equation}
    \bm{M}_{i,N_B+1} = \bm{M}_{N_A+1, j} = \bm{M}_{N_A+1,N_B+1} = r \in \mathbb{R}
\end{equation}
where $r$ is a learning parameter. 
Given a threshold $\alpha$, $i$ and $j$ with ${\bm M_{i,j}>\alpha}$ are selected as initial correspondence, where $j=\max_k{\bm M_{i,k}}$ and $i=\max_k{\bm M_{k,j}}$.

\begin{algorithm}[!t]
    \caption{Self-attention based compatibility estimation.}
    \label{alg:1}
    \begin{algorithmic}[1]
        \Require Initial correspondences ${\bm f}_\text{in}\in\mathbb{R}^{N\times4}$\,;
        \Ensure Correspondence features ${\bm f}_\text{out}\in\mathbb{R}^{N\times128}$\,;
        \State Project ${\bm f}_\text{in}$ to $\bm f\in\mathbb{R}^{N\times128}$ using MLP\,; 
        \State Normalize $\bm f$ by ${\bm f}' = {\bm f}/{||{\bm f}||}_2$\,;
        \State Generate the adjacency matrix ${\bm A}={{\bm f}^{'T}}{\bm f'}$\,;
        \State Estimate the degree matrix ${\bm D}_{ii}=\sum_{j}{\bm A}_{ij}$\,;
        \State Generate ${\bm f}_{c}$ by concatenating ${\bm f}_\text{in}$ and ${\bm D}$\,;
        \State Perform \emph{KNN} and \emph{EdgeConv}~\citep{wang2019dynamic} on ${\bm f}_{c}$\,;
        \State Generate ${\bm f}_\text{out}$ by aggregating ${\bm f}_{c}$ using max pooling\,;
        \State Return ${\bm f}_\text{out}$\,;
    \end{algorithmic}
\end{algorithm}

\subsection{Fine-level Matching}
Since it is nontrivial to find reliable correspondences, outliers rejection is often executed in an iterative manner. We bypass this issue by alternatively estimating weights that indicate the likelihood of correspondences being correct and using weights as a compatibility metric to search for compatible neighbors. Specifically, we implement fine matching based on a modified neighbors mining block (NM-block)~\citep{zhao2019nm} and an attentive residual block (AR-block) that replaces batch normalization with attentive context normalization~\citep{sun2020acne} in the standard residual block~\citep{he2016deep}, as shown in Fig.~\ref{Fig:network}(b). 

The modified NM-Block implements self-attention-based compatibility estimation to obtain local features of the correspondences, which are summarized in Algorithm~\ref{alg:1}. First, we calculate the adjacency matrix ${\bf A}$ and the corresponding degree matrix ${\bf D}$ with self-attention. The diagonal element in ${\bf D}$ indicates the compatibility between the corresponding correspondence and the remaining correspondences, so correspondences with higher ${d}$ are supposed to be compatible. 

To extract local information, we perform the K-Nearest-Neighbor (KNN) ($k = 8$) on the concatenated feature map and then use joint \emph{EdgeConv}~\citep{wang2019dynamic} and max pooling.
Concretely, given a $d$-dimensional feature set with $P$ correspondences, denoted by ${\bf{T}}={\{\bf{t}}_1, {\bf{t}}_2,..., {\bf{t}}_P\} \in \mathbb{R}^{{P} \times {d}}$, we construct a directed graph $\mathcal{G}=(\mathcal{V}, \mathcal{E})$ as the $k$-nearest neighbor (KNN) graph of $\bf T$ in $\mathbb{R}^d$, where $\mathcal{V} = \{ 1, ..., P\}$ and $\mathcal{E} \subseteq \mathcal{V} \times k$ are the vertices and edges, respectively. The graph does not include self-loop, meaning that each node does not point to itself. We concatenate the feature of each vertex with those of neighboring vertices via edges, obtaining a new feature map ${\bf{T}^{'}} \in \mathbb{R}^{{P} \times {2d} \times k}$, where $k$ denotes the number of neighbors of the KNN. We define the edge feature as ${\bm e}_{ij} = h_{\Theta} ({\bf{t}}_i, {\bf{t}}_j)$, where $h_{\Theta}:\mathbb{R}^{2d} \rightarrow \mathbb{R}^{d^{'}}$ is a nonlinear function with a set of learnable parameters $\Theta$.  Then we can obtain the edge feature map ${\bf E} \in \mathbb{R}^{{P} \times d^{'} \times k}$ and apply a channel-wise max-pooling on the edge features associated with each vertex to obtain the new feature set ${\bf \hat{T}} \in \mathbb{R}^{{P} \times d^{'}}$.

Next, several AR-Blocks are followed. According to~\cite{sun2020acne}, attentive context normalization is beneficial in improving correspondence consistency. Formally, for a feature map $\bm{F} \in \mathbb{R}^{N \times H}$ ($H$ is the number of feature dimensionality), we have the following.
\begin{equation}
    \begin{split}
        \bm{Z} &= \bm{V}\bm{F}^T + \bm{b},~
        \bm{w}_i^\text{local} = \frac{1}{1 + e^{-\bm{z}_i}},~
        \bm{w}_i^\text{global} = \frac{e^{\bm{z}_i}}{\sum_{i=1}^{N} e^{\bm{z}_i}}
    \end{split}
    \,,
\end{equation}
where $\bm{V} \in \mathbb{R}^{H \times H}$ and $\bm{b}$ are learnable parameters, and each row vector $\bm{z}_i \in \bm{Z}$ denotes the weight vector for the data point $i$. Then, $\bm{F}$ can be normalized to $\bm{W}^\text{local}$ and $\bm{W}^\text{global}$ so that
\begin{equation} 
    \bm{F}_\text{norm} = (\bm{F} - \mu_{\bm{w}}(\bm{F})) \oslash \sigma_{\bm{w}}(\bm{F}) \,,
\end{equation}
where $\oslash$ denotes the element-wise division, $\mu_{\bm{w}}(\bm{F}) =\mathbb{E}[ \bm{W}^\text{local} \\ \otimes \bm{W}^\text{global}\otimes\bm{F} ]$ and $\sigma_{\bm{w}}(\bm{F}) = \sqrt{\mathbb{E}\left[(\bm{F} -\mu_{\bm{w}}(\bm{F}))^2\right]}$.

\subsection{Loss Function}
We have two loss functions ${L_{{\rm coarse}}}$ and ${L_{{\rm fine}}}$ for the coarse matching stage and the fine matching stage, respectively. To supervise the assignment matrix ${\bm M}$, we generate ground truth correspondences ${\mathcal O}$ based on camera poses and depth following SuperGlue~\citep{sarlin2020superglue}. Given the predicted correspondences ${\mathcal Q}$, ${L_{{\rm coarse}}}$ can be defined by
\begin{equation}
\begin{split}
    L_{\rm coarse} &= -\sum_{(i,j)\in{\mathcal O}} \bm M_{i, j} \\
       &- \sum_{(i,j){\notin{\mathcal O}}\cap (i,j){\in{\mathcal Q}}}\left[\bm M_{i, N_B+1} + \bm M_{N_A+1, j}\right]
\end{split}\,,
\end{equation}
where $N_A$ and $N_B$ are the number of keypoints in an image pair. $L_{\rm fine}=L_{\rm bce}+\lambda L_{\rm geo}$ is a hybrid loss function consisting of a binary cross-entropy loss ${L_{\rm bce}}$ and a geometry loss ${L_{\rm geo}}$ based on the epipolar distance~\citep{andrew2001multiple}, defined by
\begin{equation}\small
     L_{\rm geo} = \tt  \frac{(p_2^T\hat{E}p_1)^2}{{||{Ep_1}||}_{[1]}^2 + {||{Ep_1}||}_{[2]}^2 +{||{E^{T}p_2}||}_{[1]}^2 + {||{E^{T}p_2}||}_{[2]}^2}\,,
\end{equation}
where ${\tt p_1}$ and ${\tt p_2}$ are correspondences, ${\tt\hat{E}}$ is the essential matrix predicted by the eight-point algorithm~\citep{andrew2001multiple}, and ${\bm h_{[i]}}$ denotes the ${i}$ th element of the vector ${\bm h}$.


\section{DEGREE: Encoding Correspondence Coordinates}
\label{sec:c2c}

Given a few correspondences and the corresponding confidence values generated by SMFNet, we then sort the correspondences in descending order conditioned on the confidence values, and the top-ranked correspondences are chosen as reliable seed correspondences (anchors). Using these anchors, we show how to dynamically encode the coordinates to improve the feature correspondences. We first introduce a barycentric coordinate system and demonstrate its geometrically invariant properties. Then we study correspondence coordinates in sparse feature matching and consistency filtering. We call this new approach Dynamic Encoding in the Geometric-invaRiant coordinatE systEm (\ourmethod).

\subsection{Barycentric Coordinates Revisited}

\begin{figure}[!t]
    \subfigure[]{
    \includegraphics[width=0.25\textwidth]{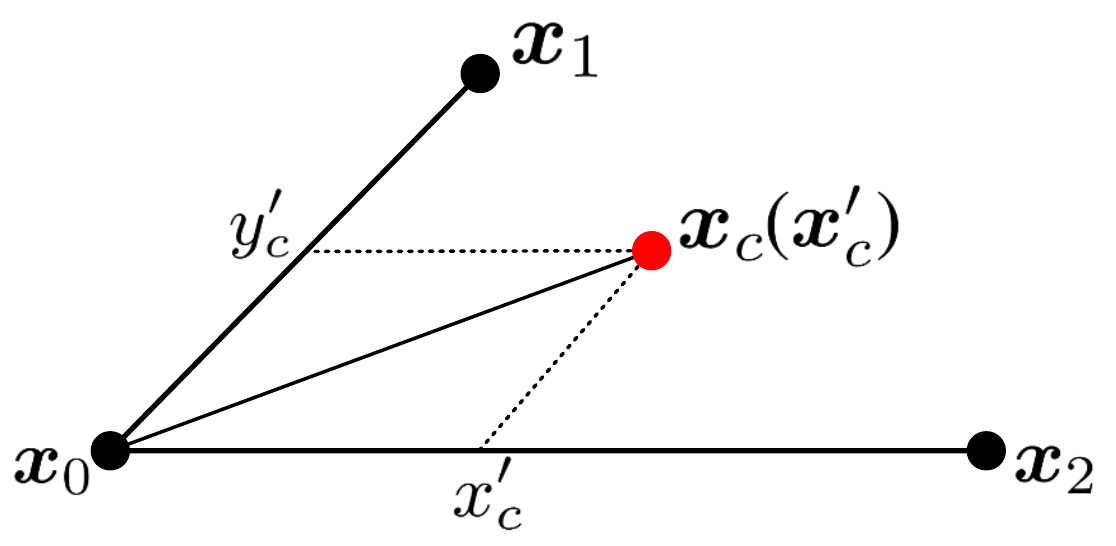}}
    \subfigure[]{
    \includegraphics[width=0.20\textwidth]{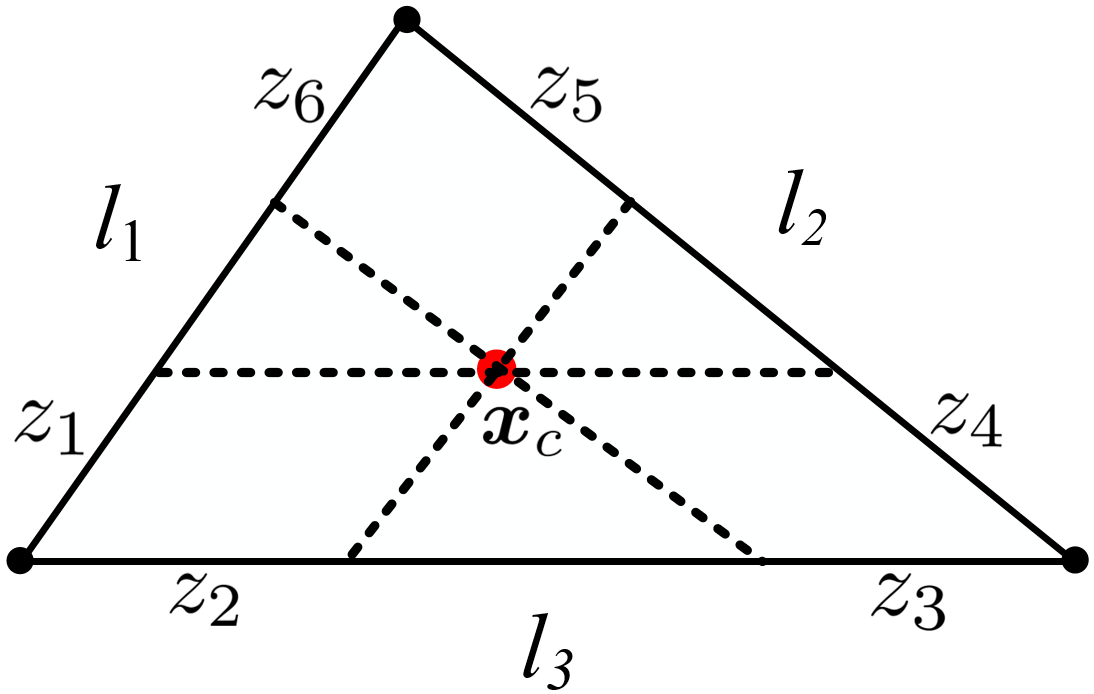}}
   \caption{(a) Barycentric coordinate system. The system is built by $3$ points ${\bm x}_0$, ${\bm x}_1$, and ${\bm x}_2$. ${\bm x}_c$ is projected to ${\bm x}_c'=\{x_c', y_c'\}$ to obtain the barycentric coordinate. (b) Geometric interpretation of the barycentric coordinate. ${\bm x}_c$ is a point to be projected, $l_i$ is the length of the axis, and ${z}_i$ is the projection length of ${\bm x}_c$ on different axes.} 
\label{fig:bary_coord}
\end{figure}

The Cartesian coordinate system is a straightforward representation of position. To address difficult cases like repeated patterns in correspondences, a key insight of this work is that one should encode geometric-invariant coordinates, such as barycentric coordinates. The barycentric coordinate system~\citep{marschner2018fundamentals} has been widely used in graphics due to its geometric invariance under rigid transforms. The standard formulation of barycentric coordinates is defined by the ratio of areas in a triangle. For an intuitive understanding of the barycentric coordinate system, here we provide an alternative interpretation of barycentric coordinates from the perspective of the {\em vector projection}.

\vspace{10pt}
\noindent\textbf{Barycentric Coordinates}.
The barycentric coordinate system is built conditionally on a set of geometric bases. Fig.~\ref{fig:bary_coord}(a) shows a set of bases with three basis vectors, \ie, ${\bm{x}}_i=(x_i, y_i)$, $i=0,1,2$, where ${\bm x}_0$ is the origin and the vector space spanned by ${\bm x}_1$ and ${\bm x}_2$ is defined by
\begin{equation}
\begin{cases}
\overrightarrow{{\bm x}_{0}{\bm x}_{1}}=(x_1-x_0, y_1-y_0)^{T}=(\Delta x_1,\Delta y_1)^{T}\\
\overrightarrow{{\bm x}_{0}{\bm x}_{2}}=(x_2-x_0, y_2-y_0)^{T}=(\Delta x_2, \Delta y_2)^{T}
\end{cases}\,.
\end{equation}
According to the basic principle of the coordinate system, $\overrightarrow{{\bm x}_{0}{\bm x}_{1}}$ and $\overrightarrow{{\bm x}_{0}{\bm x}_{2}}$ cannot be parallel. Therefore, we have the following constraint.
\begin{equation}
\overrightarrow{{\bm x}_{0}{\bm x}_{1}}\circ\overrightarrow{{\bm x}_{0}{\bm x}_{2}}=\Delta x_1\Delta y_2-\Delta y_1\Delta x_2\neq0\,,
\end{equation}
where $\circ$ denotes the inner product. If the above condition is not satisfied, another point will be chosen as the origin of the alternative candidates. Any point
${\bm x}_c$ can be projected onto the basis vectors using the new basis set ${\mathcal B}=\{\overrightarrow{{\bm x}_{0}{\bm x}_{1}},\overrightarrow{{\bm x}_{0}{\bm x}_{2}}$\} with a transition matrix as follows (note that the basis vectors are not normalized). 

In vector space, the coordinates of any vector under a specific basis are unique, but the coordinates under different bases are generally different. Therefore, a transition matrix is often used to characterize the relationship between bases. Given the natural basis $\bm e$ and the new set of bases ${\mathcal{B}}$, the transition matrix ${\bm T}$ can be written as
\begin{equation} \label{T_expression}
\bm T=
\begin{bmatrix}
\Delta x_1 & \Delta x_2 \\
\Delta y_1 & \Delta y_2
\end{bmatrix}\,.
\end{equation}

To calculate the new coordinates of the projected points,
we denote the coordinate of ${\bm x}_c$ under the natural basis $\bm e$ by \mbox{$(x_c-x_0, y_c-y_0)^{T}$}, and the transited coordinate under ${\mathcal B}$  by $(x_c', y_c')^{T}$. Then, we have
\begin{equation} \label{transfer3}
  {\bm e}\begin{bmatrix}\begin{array}{c}\Delta x_c\\ \Delta y_c\end{array}\end{bmatrix} = {\bm T}\begin{bmatrix}\begin{array}{c}x_c'\\ y_c'\end{array}\end{bmatrix}\,,
\end{equation}
where
\begin{equation} \label{transfer4}
   {\bm e} =\begin{bmatrix}1 & 0 \\0 & 1 \end{bmatrix}\,,\,\,
   \begin{bmatrix}\begin{array}{c}\Delta x_c\\ \Delta y_c\end{array}\end{bmatrix} = \begin{bmatrix}\begin{array}{c}x_c-x_0\\ y_c-y_0\end{array}\end{bmatrix} \,.
\end{equation}
It follows from Eq.~\eqref{transfer3} that the coordinate of ${\bm x}_c$ under ${\mathcal B}$ can be obtained by
\begin{equation} \label{x_expression}
  \begin{split}
  {\bm x}_c' &= \begin{bmatrix}\begin{array}{c}x_c'\\ y_c'\end{array}\end{bmatrix} ={\bm T}^{-1} \begin{bmatrix}\begin{array}{c}\Delta x_c\\ \Delta y_c\end{array}\end{bmatrix}
  = \rho \begin{bmatrix}\begin{array}{c} \Delta y_2\Delta x_c-\Delta x_2\Delta y_c\ \\
    -\Delta y_1\Delta x_c+
    \Delta x_1\Delta y_c\
    \end{array}\end{bmatrix}
   \\
  \end{split}\,,
\end{equation}
where
\begin{equation} \label{eq:c}
    \rho = \frac{1}{\mid \Delta x_1\Delta y_2-\Delta x_2\Delta y_1 \mid} > 0\,. 
\end{equation}
Eq.~\eqref{eq:c} guarantees that the two axes of the new coordinate system are not parallel. In practice, $\bm x_i$'s are chosen from the seed correspondences discussed in Section~\ref{sec:seed choosen}. Note that here the coordinate of ${\bm x}_c$ is only a partial barycentric coordinate of a point.

To obtain the full barycentric coordinate of ${\bm x}_c$, we need to construct two additional barycentric coordinate systems from the same $3$ basis points, as shown in Fig.~\ref{fig:bary_coord}(b). According to Eq.~\eqref{x_expression}, three sets of partial barycentric coordinates can be obtained. Since the basis axes are not normalized, the ratio between the projection length, denoted by $z_i$, and the length of the basis axis, denoted by $l_i$. In this way, we have $3$ sets of coordinates $(\frac{z_1}{l_1}, \frac{z_2}{l_3})$, $(\frac{z_3}{l_3}$, $\frac{z_4}{l_2})$, and $(\frac{z_5}{l_2}, \frac{z_6}{l_1})$. According to the triangle proportionality theorem, we have the following.
\begin{equation} \label{encoding barycentric}
    \frac{z_1}{{l}_1} = \frac{z_4}{l_2}\,,
    \frac{z_2}{l_3} = \frac{z_5}{l_2}\,,
    \frac{z_3}{l_3} = \frac{z_6}{l_1}\,.
\end{equation}
Therefore, the three sets of coordinates can be simplified to the full barycentric coordinate of a point $(\frac{z_1}{l_1}, \frac{z_5}{l_2}, \frac{z_3}{l_3})$. Alternatively, the barycentric coordinate $\bm{x}_c'$ in Eq.~\eqref{x_expression} can also be represented by
\begin{equation} \label{xc_hat}
    \hat{\bm{x}}_c = \left[\bm{x}_c', 1-\parallel\bm{x}_c'\parallel _1\right]^T \,.
\end{equation}

\noindent\textbf{Geometric Invariance}.
The barycentric coordinate system has many invariant properties, including translation invariance, scale invariance, rotation invariance, and affine invariance. According to Eq.~\eqref{encoding barycentric}, barycentric coordinates are represented by projection lengths and are normalized by basis axes. It is easy to infer that the projection lengths are invariant to translation and/or rotation. Also, since barycentric coordinates are always normalized by the basis axes, the coordinates remain unchanged in scaling. Here, we prove the two important properties of invariance (rotation and affine invariance) of the geometric coordinate system.

Let $\bm{x}_i^n$ be the points that make up the base set, where $n$ indicates the $n$-th coordinate system and $i$ is the $i$-th point. Generally, $\bm{x}_0^n$ is set as the origin. $\bm{T}_n$ is the transition matrix and $\mathcal{P}_n$ represents the set of points to be projected. For $\forall \bm p_k^n \in \mathcal{P}_n$, its coordinates are denoted by $[x_k^n, y_k^n]^T$. Without loss of generality, we discuss the properties between two sets of bases formed by $\bm x_i^0$'s and $\bm x_i^1$'s.

\begin{theorem}[Rotation Invariance]
Given a rotation transformation ${f: {\bm x}_i^0 \rightarrow {\bm x}_i^1}$ with respect to a rotation matrix ${\bm R}_{2\times 2}$\, such that $ {\bm x}_i^1= {\bm R}{\bm x}_i^0 $\,. Then, for~$\forall \bm x_i^1$\,,
$ {\bm T}_1 = {\bm R}{\bm T}_0$\,.
Therefore, for~$\forall \bm p_k^n \in \mathcal{P}_n$\,, $ {{\bm p}_k^1}' = ({\bm T}_0^{-1}{\bm R}^{-1}) {{\bm p}_k^1} = {{\bm p}_k^0}'$\,.
\end{theorem}

\begin{theorem}[Affine Invariance]
Given an affine transformation ${f: \begin{bmatrix}{\bm x}_i^0 \\ 1\end{bmatrix} \rightarrow {\bm x}_i^1}$ 
w.r.t.\ a transformation matrix ${\bm A}_{2\times3}$\,, such that ${\bm x}_i^1= {\bm A}_{2\times3}\begin{bmatrix}{\bm x}_i^0 \\ 1\end{bmatrix}$\,. Then, for~$\forall \bm x_i^1$\,,
${\bm T}_1 = \hat{{\bm A}}_{2\times2}{\bm T}_0$\,,
where ${\bm A}_{2\times3} = {\begin{bmatrix}
{\hat{{\bm A}}_{2\times2}},{\bm C}_{2\times1}
\end{bmatrix}}$ and ${\bm C}_{2\times1}$ is a translation matrix. Therefore, for~$\forall \bm p_k^n \in \mathcal{P}_n$\,, $ {{\bm p}_k^1}' = ({\bm T}_0^{-1}\hat{{\bm A}}_{2\times2}^{-1}) {{\bm p}_k^1} = {{\bm p}_k^0}'$\,.
\end{theorem}
Rigorous proofs of these two theorems can be found in the appendix.

\begin{figure}[!t]
    \centering
    \subfigure[]{
    \includegraphics[width=0.45\textwidth]{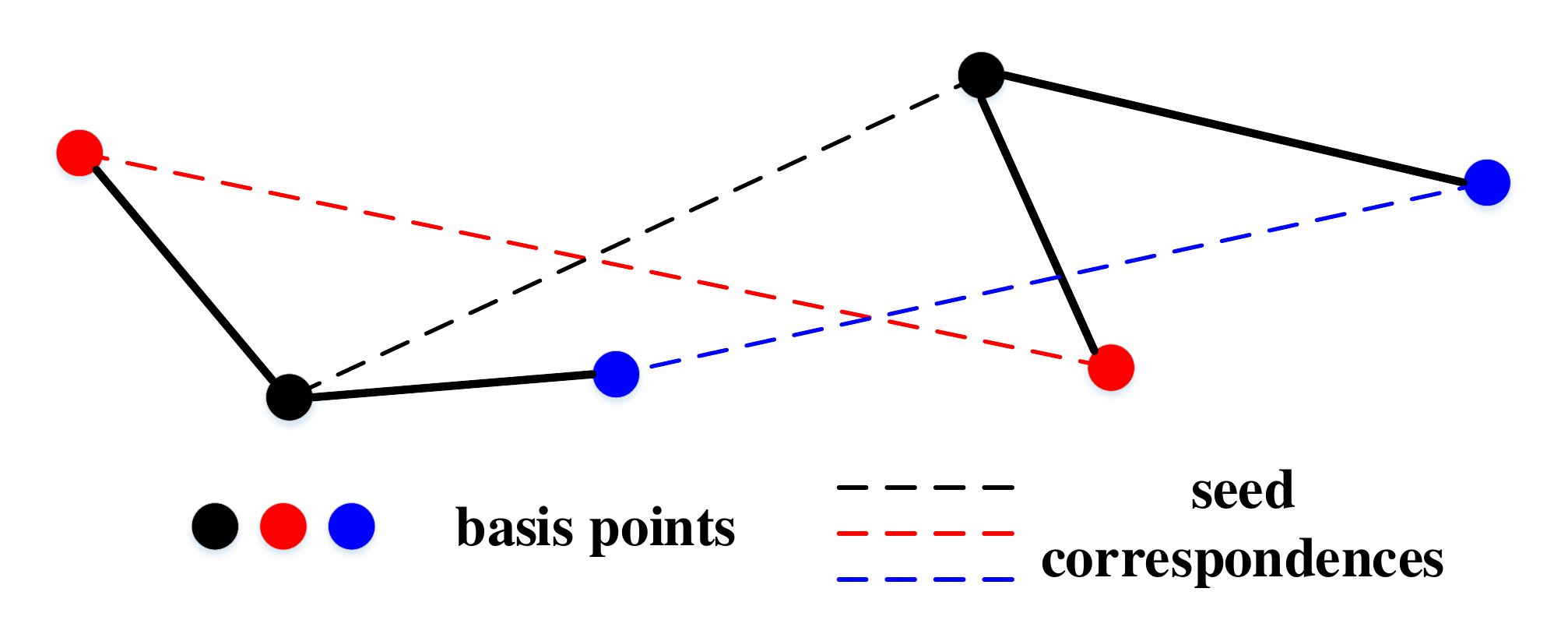}}
    \subfigure[]{
    \includegraphics[width=0.40\textwidth]{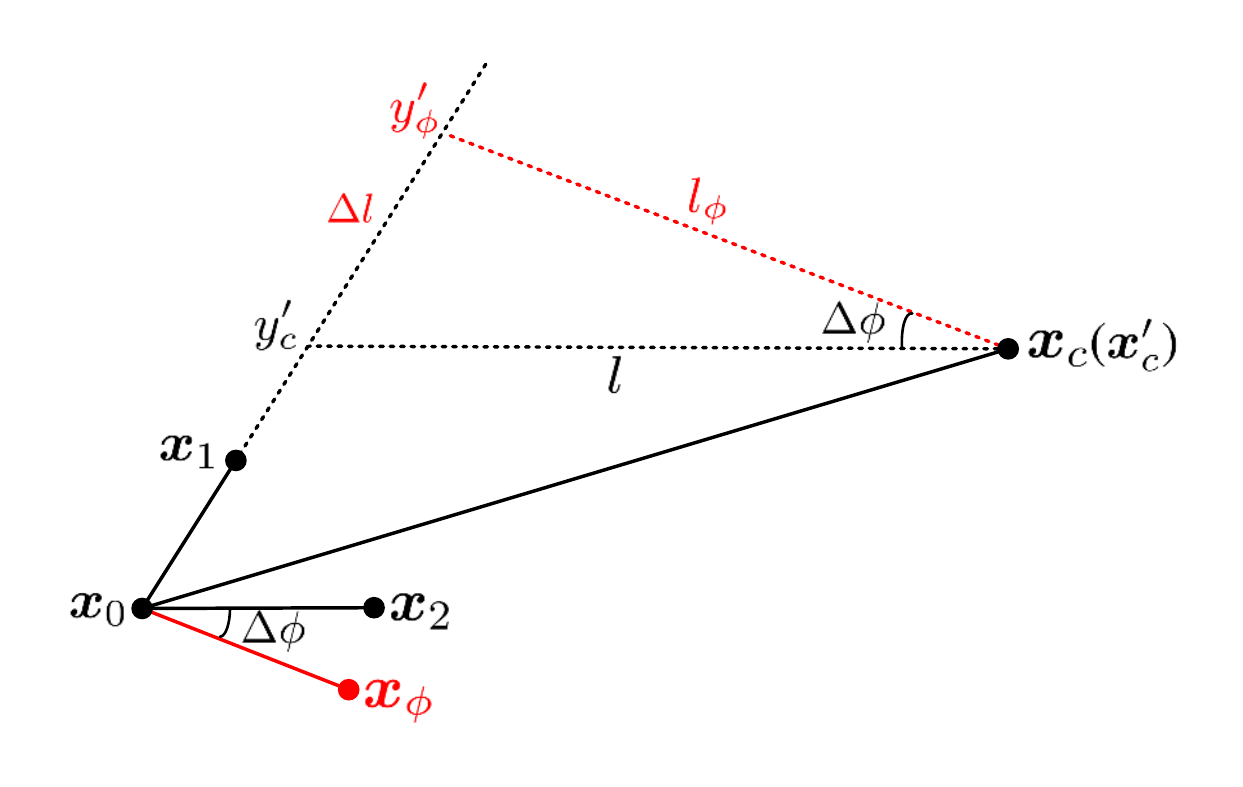}}
    \caption{(a) Paired barycentric coordinate systems. The two coordinate systems are formed by three seed correspondences. Different colors represent different correspondences. (b) Biased barycentric coordinates with shifted correspondences. ${\bm x}_c$ is a point to be projected. {${\bm x}_\phi$} represents $\bm{x}_2$ with disturbance $\Delta\phi$. {$y_\phi'$} and $y_c'$ represent the coordinate alone $\bm{x_0 x_1}$ axis under different coordinate systems. $\Delta l$ denotes the absolute value of the difference between {$y_\phi'$} and $y_c'$. {$l_\phi$} and $l$ denote the distances, between the point ${\bm x}_c$ and the axis $\bm{x_0 x_1}$, along $\bm{x_0 {x}_\phi}$ and $\bm{x_0 x_2}$ direction respectively.}
\label{pic:paired correspendence}
\end{figure}

\subsection{Barycentric Correspondence Coordinates}

\begin{figure}[!t]
    \centering
    \includegraphics[width=0.95\linewidth]{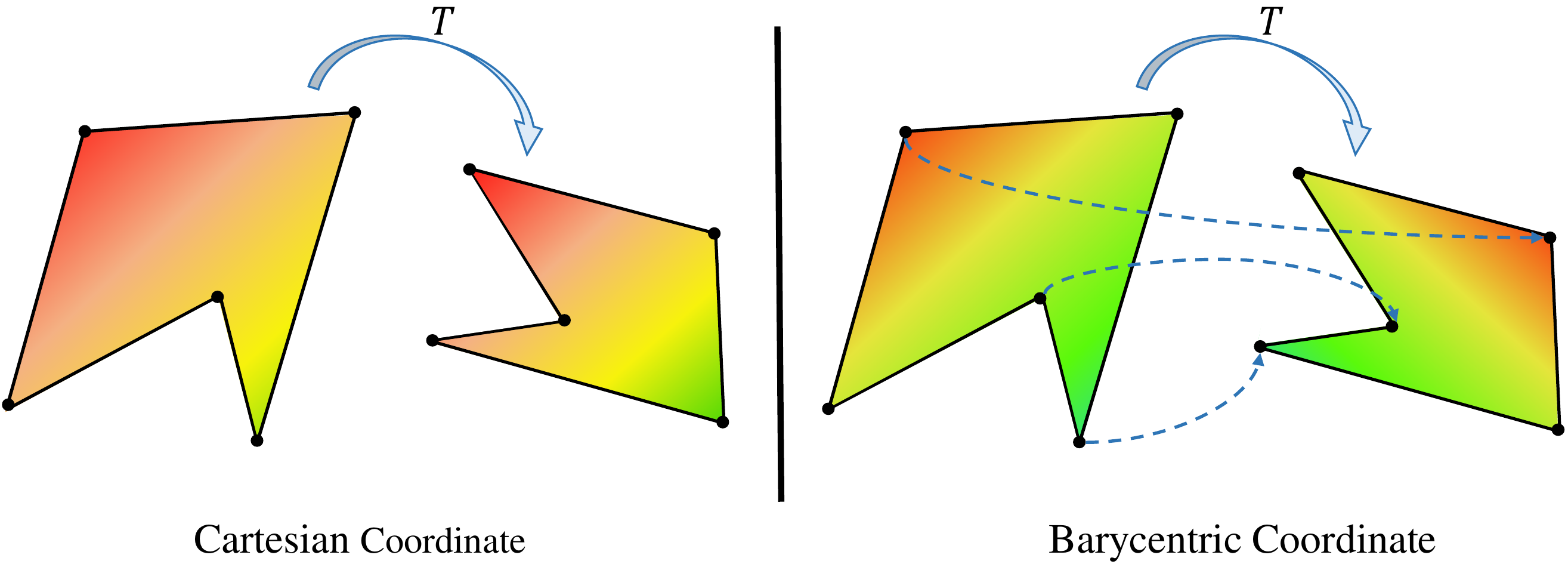}
    \caption{Comparison between Cartesian and barycentric coordinates. With an affine transformation matrix $T$, we encoder different coordinates into RGB channel by the same color map. Apparently, the barycentric coordinate ensures the invariance of the coordinate representation under affine transformation, which benefits from its affine invariant properties. }
\label{pic:coordinate color}
\end{figure}

In the context of feature correspondence, where image pairs are considered, we show how to generate barycentric correspondence coordinates from a pair of barycentric coordinate systems. In particular, we assign each keypoint in an image pair a coordinate representation. To increase the robustness of the barycentric coordinate system, we propose the selection of multiple basis points from various seed correspondences. In this way, for every image, we have a set of barycentric coordinate systems; each one is constructed by three basis points. Each basis point of an image is associated with a seed correspondence and has a matched pair of basis points in the given image pair, as shown in Fig.~\ref{pic:paired correspendence}(a). By sequentially projecting each point to each system, we can obtain correspondence coordinates of a point by concatenating all barycentric coordinates from different coordinate systems. Note that each point of an image is projected only onto the barycentric coordinate systems specific to this image. We visually compare the affine invariance of barycentric coordinates with Cartesian coordinates, given an affine transformation image pair shown in Fig.~\ref{pic:coordinate color}. 

In practice, we can integrate the correspondence coordinates with existing descriptors/features to enhance their sensitivity to position. The integrated descriptors/features can be used as usual with feature matchers and/or consistency filters. Suppose that $K$ basis points are available ($K\geq3$) in each image and that we can build $N(K)=K\cdot(K-1)\cdot(K-2)/6$ barycentric coordinate systems. Therefore, the final coordinate representation of a point at the image level is of $N(K)$ dimensionality.

\begin{figure*}[!t]
    \centering
    \includegraphics[width=1.0\textwidth]{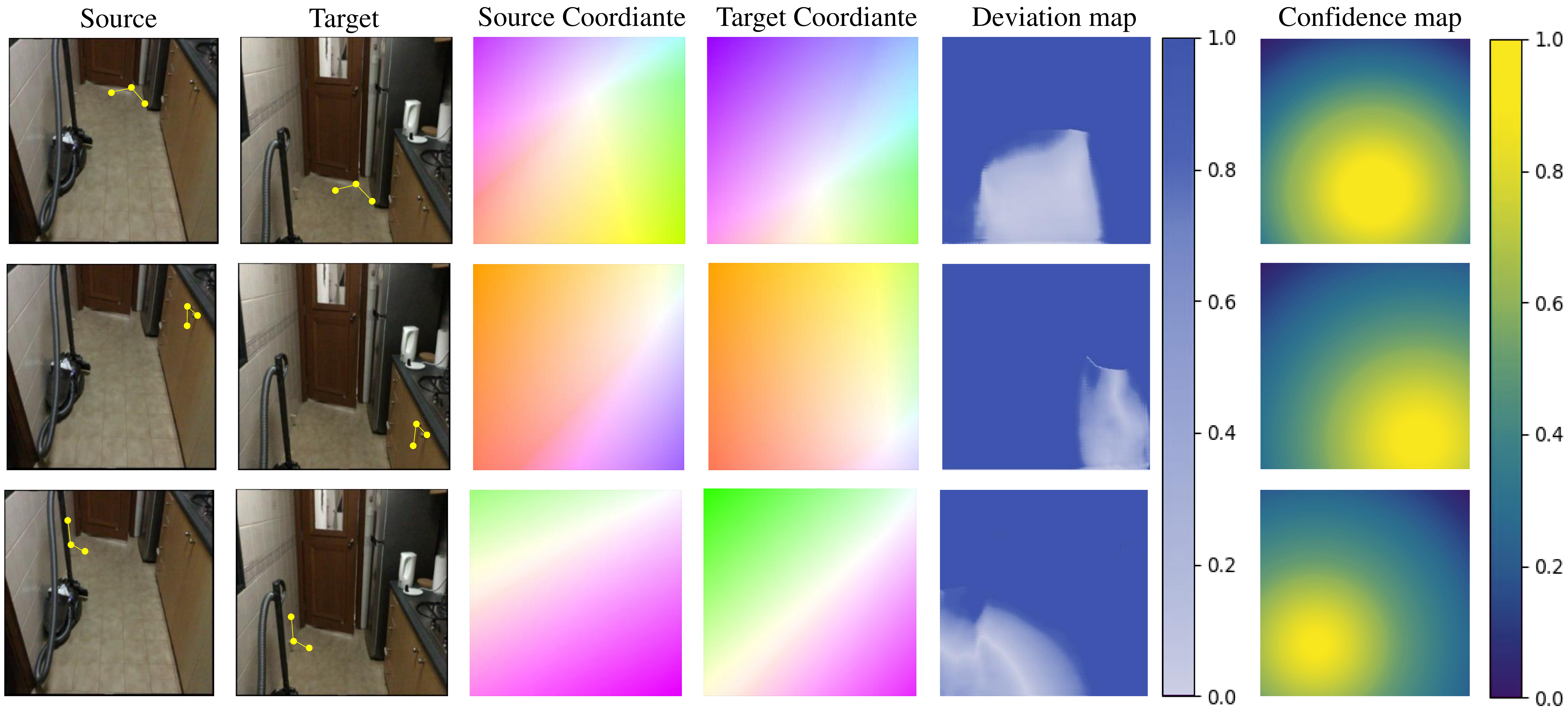}
    \caption{Multiple barycentric correspondence coordinates from an image pair in the indoor scene. The barycentric coordinate maps of the source and the target image are color-coded in different coordinate systems, respectively. The deviation maps that characterize the coordinate differences show that the encoded barycentric coordinates become less accurate in regions far away from the coordinate origin. To alleviate this, confidence maps can be used to inform inaccurate regions.}
\label{pic:indoor_error}
\end{figure*}

Critically, the strategy of multiple barycentric coordinate systems breaks the original 2D affine invariant constraint of a single barycentric coordinate system. Different barycentric coordinate systems decompose the real-world 3D transformation into a series of 2D image transformations in different planes. This strategy encourages \ourmethod to better fit the realistic camera pose transformation. As shown in Fig.~\ref{pic:indoor_error}, the left wall, middle floor, and right cabinet in this scene belong to three different planes. \ourmethod can establish multiple barycentric coordinate systems on different planes to fit the overall geometric transformation of the entire scene.

One may be concerned that different coordinate representations generated by \ourmethod cannot be compared due to the dynamic nature of seed correspondences. We remark that this is not a problem for \ourmethod. Due to the invariance properties of barycentric coordinates, different coordinates have consistent representations, even if they are generated by varying coordinate systems.

\vspace{10pt}
\noindent\textbf{When Geometric Invariance Holds}. 
To safely use the correspondence coordinates, it is important to know when the property of geometric invariance holds. Since barycentric coordinate systems are established using three pairs of seed correspondences in the image, the geometric invariance property of established barycentric coordinate systems holds if and only if the following conditions are satisfied:

\begin{remark} \label{remark-1}
The three pairs of seed correspondences that build the basis points follow a 2D affine transformation.
\end{remark}

\begin{remark} \label{remark-2}
The keypoint to be represented lies on the same plane as the three base points.
\end{remark}

\begin{remark} \label{remark-3}
The seed correspondences are accurate.
\end{remark}

\begin{figure}[!t]
    \centering
    \includegraphics[width=0.48\textwidth]{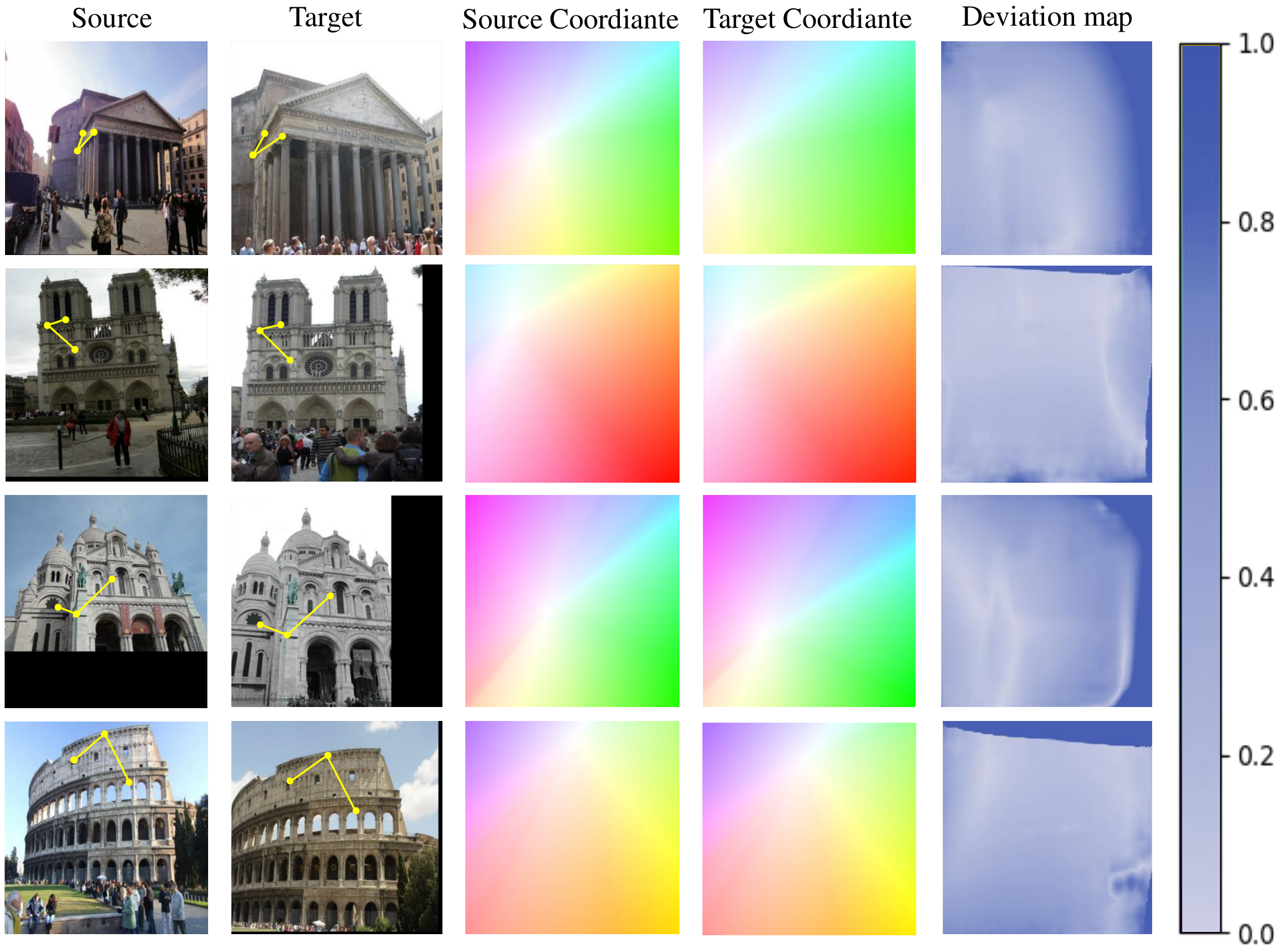}
    \caption{The barycentric coordinate maps in outdoor scenes. We color-code the barycentric coordinate maps of the source and the target image. A deviation map is also computed by comparing the coordinate difference between the target coordinate map and the ground-truth coordinate map, where the ground-truth coordinate map is obtained by remapping the target coordinate map using the ground-truth flow map. Each coordinate deviation value $d$ is clipped by $\min(1,d)$ for better visualization. The brighter regions show in the deviation map, the more consistent the barycentric coordinates between the image pair are in such regions.
    }
\label{pic:outdoor_error}
\end{figure}

Condition~\ref{remark-3} could hold to a large extent using our proposed SMFNet, but Condition~\ref{remark-1} and Condition~\ref{remark-2} rarely hold from a theoretical point of view in real-world circumstances, with complex $3$-D structures and $3$-D camera transformations. 
However, from a practical point of view, such preconditions can be approximately met under certain scenarios. In particular, given an object in the $3$-D space, when the distance between the camera and the object is much larger than the depth variations of the object, such depth variations can be ignored. Thus, the keypoints extracted from the object can be considered to be on the same plane. Therefore, the geometric transformation $3$ -D is reduced to an affine or homography transformation in the image $2$ -D. 
The scenarios above often appear in outdoor scenes. From Fig.~\ref{pic:outdoor_error}, we select image pairs from the PhotoTourism dataset, construct barycentric coordinate systems using the basis keypoints, visualize the color-coded barycentric coordinate maps, and compute a deviation map to indicate the coordinate difference between the target coordinate map and the ground-truth coordinate map. The deviation map shows that the barycentric correspondence coordinates possess geometric invariance to a great degree and generally fit the geometric transformation in outdoor scenes.
However, in some scenarios, particularly indoor scenes, the assumption of the approximate $2$-D image transformation does not hold due to the limited object-camera distance and the large variations of depth of field, resulting in unreliable correspondence coordinates shown in Fig.~\ref{pic:indoor_error}, particularly when the keypoints are far from the origin. To address this, we introduce a solution employing multiple barycentric coordinate systems along with correspondence confidence maps.

\subsection{Correspondence Coordinate Confidence}

To tackle the problem of inaccurate correspondence coordinates, we propose to use a confidence map to identify the inaccurate coordinate encoding regions and filter out these coordinates, explained in the following.

Intuitively, the position of each seed point cannot be completely accurate, especially for sparse matching; there may be a deviation of several pixels. If an axis of the barycentric coordinate system has a small deviation, as shown in Fig.~\ref{pic:paired correspendence}(b), $\bm{x_0 x_2}$ would have an angle bias of $\Delta\phi$ and thus lead to a coordinate deviation.
\begin{equation} \label{eq:delta_l}
\Delta l = \mid y_{\phi}' - y_c' \mid \,.
\end{equation}
When $\bm{x}_c$ is far from $\bm{x}_0$, we have $l_\phi \approx l$ (see Fig.~\ref{pic:paired correspendence}(b)). Due to the tiny deviation of $\bm{x}_2$, we also have $\Delta\phi \rightarrow 0$. Then, Eq.~\eqref{eq:delta_l} can be written as,
\begin{equation} 
\Delta l \approx l \cdot \Delta\phi \,.
\end{equation}
For example, given an image of size $1280 \times 960$, if $\Delta\phi = 4^\circ$ and $ l = 150$, $\Delta l\approx11$. With increasing $\Delta\phi$ and $l$, $\Delta l$ will also increase and cannot be ignored. Therefore, we propose to assign a confidence weight $\omega_c$ for each keypoint coordinate, \ie,
\begin{equation} \label{eq:weight}
\omega_c = e^{-\frac{1}{\varepsilon}\frac{\parallel\overrightarrow{{\bm x}_{0}{\bm x}_{c}}\parallel_2}{ \parallel\overrightarrow{{\bm x}_{0}{\bm x}_{1}} + \overrightarrow{{\bm x}_{0}{\bm x}_{2}}\parallel_2}} \,.
\end{equation}
$\omega_c$ gradually decreases as the distance between $\bm{x}_c$ and $\bm{x}_0$ increases, indicating that the confidence in the coordinates gradually decreases. $\varepsilon$ in Eq.~\eqref{eq:weight} controls the decreasing speed of $\omega_c$.
Next, we will introduce how to use correspondence coordinates in sparse feature matching and consistency filtering.

\subsection{Correspondence Coordinates in Sparse Feature Matching and Consistency Filtering} \label{sec:4-3}

\noindent\textbf{Sparse Feature Matching.}
To integrate position information with local feature descriptors, keypoint coordinates are often encoded by an encoding network in sparse feature matching. Here, we explain how to use the correspondence coordinates in this context. 

In particular, our idea is to feed the normalized and confidence-weighted correspondence coordinates into an encoding network to generate geometric information between an image pair. Given an image pair $I^A$ and $I^B$, we can generate 
the correspondence coordinates $\bm{P}^A \in \mathbb{R}^{N\times3D}$ and $\bm{P}^B\in \mathbb{R}^{M\times3D}$, respectively, with $K$ seed correspondences, where $N$ and $M$ are the number of key points in $I^A$ and $I^B$, respectively. 
If the $i^\text{th}$ row of $\bm{P}^\star$ is denoted by $\bm{p}_i^\star$ ($\star=A,B$), $\bm{p}_i^\star$ has the form $\bm{p}_i^\star = \{ \hat{\bm{x}}_{i(1)}^\star, ~\dots \,,\hat{\bm{x}}_{i(d)}^\star\,,~\dots\,, \hat{\bm{x}}_{i(D)}^\star \}$, where $ D =N(K)$, and $\hat{\bm{x}}_{i(d)}^\star$ denotes the barycentric coordinate of the $d^\text{th}$ coordinate system in the image $\star$. 
In the open literature~\citep{zhang2019learning,sarlin2020superglue}, zero-score normalization is often applied to input coordinates. Here, we apply zero-score normalization along the first dimension of $\bm{P}$, where $\bm{P} = {\tt{cat}}({\bm{P}}^A, \,{\bm{P}}^B) \in \mathbb{R}^{(N+M)\times3D}$. 
Let $\Tilde{\bm{p}}_i^\star$ be the normalized coordinates of $\bm{p}_i^\star$. 
We further weight $\Tilde{\bm{p}}_i^\star$ by its confidence according to Eq.~\eqref{eq:weight}
\begin{equation} \label{eq:superglue}
    \bm{p}_{\omega i}^\star =  \Tilde{\bm{p}}_i^\star \otimes \bm{\omega}_i^\star  \,,
\end{equation}
where $\bm{\omega}_i = [\omega_{i(1)}, \,\omega_{i(2)}, \,\dots, \,\omega_{i(D)}]$\,, and $\otimes$ denotes element-wise multiplication. Specifically, $\omega_{i(D)} \in \bm{\omega}_i$ is the weight described in Eq.~\eqref{eq:weight}, corresponding to the $D$-th barycentric coordinate system.

\vspace{5pt}
\noindent\textbf{Consistency Filtering.}
The standard input for consistency filtering is a concatenated four-dimensional coordinate vector for each correspondence. Intuitively 
such a coordinate vector would lack global position information. Therefore, we propose to use the 
correspondence coordinates to generate a two-dimensional vector and concatenate it to the 
input 
to add global geometric consistency information. Next, we show how to compute this $2$-dimensional vector.

Given a set of prefiltered correspondences $\bm{C} \in \mathbb{R}^{N\times4}$, we first encode each correspondence $\bm{c}_i\in\bm{C}$ with $K$ seed correspondences using a transformation $\Omega: \mathbb{R}^{4}\rightarrow\mathbb{R}^{6D}$, \ie, 
\begin{equation}
        \Omega(\bm{c}_i) = \left[\Omega(\bm{x}_i^A), \,\Omega(\bm{x}_i^B)\right] = [{\bm{p}}_i^A, \,{\bm{p}}_i^B]
        \,,
\end{equation}
where $\bm{x}_i^A \in \mathbb{R}^{2}$ and $\bm{x}_i^B \in \mathbb{R}^{2}$ represent the Cartesian coordinates of a pair of keypoints for every correspondence, and ${\bm{p}}_i^A \in \mathbb{R}^{3D}$ and ${\bm{p}}_i^B \in \mathbb{R}^{3D}$ are encoded correspondence coordinates. We then compute the difference of the two vector sets (${\bm{p}}_i^A$ and ${\bm{p}}_i^B$) 
and compute the sum of squares for all three dimensions to amplify the differences, characterized by a transformation $\mathcal{F}: \mathbb{R}^{3D}\rightarrow\mathbb{R}^{D}$, \ie,
\begin{equation}
    \begin{split}
        \bm{u}_{i} &= \mathcal{F}({\bm{p}}_i^A - \bm{p}_i^B) \\
        &=\left[ \parallel\hat{\bm{x}}_{i(1)}^{A} - \hat{\bm{x}}_{i(1)}^{B}\parallel_2^2, \,\dots ,\,\parallel\hat{\bm{x}}_{i(D)}^{A} - \hat{\bm{x}}_{i(D)}^{B}\parallel_2^2\right] \,,
    \end{split}
\end{equation}
where $\hat{\bm{x}}_{i(D)} \in \mathbb{R}^3$ denotes the barycentric coordinate representation in Eq.~\eqref{xc_hat}.
Then, combining $\bm{u}_i$ with the weight vector $\bm{\omega}_i$, we have the following.
\begin{equation}
    \bm{u}_{\omega i} = \bm{u}_i \otimes \bm{\omega}_i^A \otimes \bm{\omega}_i^B \,,
\end{equation}
where $\bm{\omega}_i$ is defined in Eq.~\eqref{eq:superglue}, and $\otimes$ denotes element-wise multiplication.
Finally, a two-dimensional feature vector can be obtained by computing the mean and variance of the $D$-dimensional vector, 
\ie,
\begin{equation} \label{eq:oanet}
    m_i = {\mathbb{E}}(\bm{u}_{\omega i}), \,
    v_i = \sqrt{\mathbb{E}\left[(\bm{u}_{\omega i} - {\mathbb{E}}(\bm{u}_{\omega i}))^2\right]} \,.
\end{equation}

\section{Experimental Results and Discussions}
\label{sec:cfc}

In this section, we report on our experimental results to demonstrate the performance of DEGREE. First, we design a synthesized experiment on a toy dataset to highlight the strength of \ourmethod to address the problem with repeated patterns. We then show how \ourmethod works with various feature matching and consistency filtering methods to achieve state-of-the-art performance. We have also conducted ablation studies to analyze the role of SMFNet in \ourmethod and to verify its reliability by applying it to pose estimation. Finally, we report our results at the CVPR 2021 Image Matching Challenge, where \ourmethod plays a key role in our entry into the competition.

\subsection{Baselines, Datasets, and Protocols} 

\vspace{5pt} \noindent\textbf{Baselines.}
To demonstrate the superiority and generality of \ourmethod for robust feature correspondences, we built baselines using the state-of-the-art image matching network SuperGlue~\citep{sarlin2020superglue}, the state-of-the-art consistency filtering network OANet~\citep{zhang2019learning}, and three representative descriptors: RootSIFT~\citep{arandjelovic2012three}, HardNet~\citep{mishchuk2017working}, and SuperPoint~\citep{detone2018superpoint}. RootSIFT is a classic handcrafted descriptor with few global information via multiscale pyramids; HardNet is a patch-based descriptor with only local information; SuperPoint is a state-of-the-art learning-based descriptor with global context information. We detected up to $2$~K keypoints for all evaluated images and extracted the RootSIFT implemented by OpenCV-python. For HardNet, we use the officially released model ${\tt HardNet{++}.pth}$. For SuperPoint, we 
apply the released pre-trained model ${\tt superpoint\_v1.pth}$.

\begin{table}[!t] \small
\centering
\renewcommand{\arraystretch}{1.2}
\addtolength{\tabcolsep}{-1pt}
\caption{Reliability of seed correspondences with pre-trained SMFNet. We compare SMFNet with NN + ratio test (RT)~\citep{lowe2004distinctive}. `Top-$10$' indicates the number of correct correspondences in the top-$10$ confident candidates. `Prec.' denotes the mean matching precision under the epipolar geometry distance. `R' denotes the recall ratio of correct correspondences. }
\begin{tabular}{@{}lllccc@{}}
\toprule
Dataset            & Descriptor   & Matcher    & Top-10 & Prec.($\%$) & R($\%$) \\ \midrule
\multirow{6}{*}{PhotoTour.} & \multirow{2}{*}{RootSIFT} & NN+RT   & 5.89       & 0.696   & 0.651    \\
                                                   ~ & ~ & SMFNet   & \textbf{9.87}  & \textbf{0.969}     & \textbf{0.908} \\  \cmidrule(l){3-6} 
                             & \multirow{2}{*}{HardNet}  & NN+RT    &  6.06   &  0.714  & 0.673   \\
                                                   ~ & ~ & SMFNet   & \textbf{9.88}   & \textbf{0.944}     & \textbf{0.864} \\  \cmidrule(l){3-6} 
                             & \multirow{2}{*}{SuperPoint} & NN+RT    & 5.56  &  0.653 &  0.668 \\
                                                   ~ & ~ & SMFNet   & \textbf{9.73}           & \textbf{0.952}      & \textbf{0.910}      \\  \hline
\multirow{6}{*}{SUN3D}       & \multirow{2}{*}{RootSIFT} & NN+RT  &5.32       & 0.571  & 0.635    \\
                                                   ~ & ~ & SMFNet  & \textbf{8.82}  & \textbf{0.808} & \textbf{0.871} \\ \cmidrule(l){3-6} 
                             & \multirow{2}{*}{HardNet}  & NN+RT & 5.58      & 0.567  & 0.592    \\
                                                   ~ & ~ & SMFNet & \textbf{8.48}    & \textbf{0.776}  & \textbf{0.788} \\ \cmidrule(l){3-6} 
                             & \multirow{2}{*}{SuperPoint} & NN+RT  & 5.14  &   0.538  & 0.550     \\
                                                   ~ & ~ & SMFNet & \textbf{8.16} & \textbf{0.772} & \textbf{0.715}      \\ 
\bottomrule
\end{tabular}
\label{tab:smfnet}
\end{table}

\vspace{5pt} \noindent\textbf{Datasets.} 
We conduct experiments on indoor and outdoor datasets. For the outdoor dataset, we used the Yahoo YFCC100M dataset~\citep{thomee2016yfcc100m} and the PhotoTourism dataset~\citep{jin2020image} (a subset of the YFCC100M dataset). 
YFCC100M generates $72$ sequences of tourist landmarks according to \citep{heinly2015reconstructing}. Following the experimental setting of OANet \citep{zhang2019learning}, we use $68$ sequences for training and $4$ sequences for testing. In each sequence, image pairs with overlap beyond $50\%$ are included in the dataset. $250$~K training and $4$~K testing pairs are chosen. The camera poses provided by~\cite{heinly2015reconstructing} are used to generate the ground-truth essential matrix. The PhotoTourism dataset offers ground-truth camera poses and sparse 3D models generated using SfM~\citep{wu2011visualsfm}. This dataset has $15$ scenes where $12$ is used for training and $3$ for testing. In each scene, we generate image pairs by finding the top 10 similar images for each image according to the number of common points they included. $230$~K training and $6$~K testing pairs are selected. We used camera poses and depth images to generate ground-truth correspondences.
For indoor scenes, we use the SUN3D dataset~\citep{xiao2013sun3d} and ScanNet dataset~\citep{dai2017scannet}. The SUN3D dataset is a large-scale RGB-D video database with available camera poses and object labels. Following~\cite{zhang2019learning}, the dataset is split into $239$ scenes for training and $15$ for testing. There is no spatial overlap between the training and testing datasets. We sampled videos for every $10$ frames. In each sequence, image pairs with an overlap greater than $0.35$ are selected from the dataset. We select $1$ M pairs for training and $1500$ pairs for testing. The ScanNet dataset is another widely used indoor reconstruction dataset. We train and evaluate with pairs that have an overlap ratio in the range of $[0.1, 0.8]$. Following the sampling strategy of image pairs in~\cite{d2net2019CVPR}, we select $230$ M training and $1500$ testing pairs, discarding pairs with tiny or large overlaps. All ScanNet images and depth maps are resized to $640 \times 480$ in our implementation.

\vspace{5pt} \noindent\textbf{Comparing Approaches.}
We compare our method with a heuristic pruning strategy, a ratio test and various learning-based methods, including OANet~\citep{zhang2019learning}, SuperGlue~\citep{sarlin2020superglue}, SGMNet~\citep{sgmnet}, and ClusterGNN~\citep{clustergnn}. For a fair comparison, we re-train these models using the same dataset and the official training code. Note that ClusterGNN does not provide any code or pre-trained models. Therefore, we report the results of the original paper.

\vspace{5pt} \noindent\textbf{Protocols.}
To measure performance, we use Precision (P), Recall (R), and F-measure (F) as in~\cite{lin2014bilateral} and~\cite{ma2019locality}. We define two types of precision based on different ground truths. P${_{\tt epi}}$ represents the epipolar distance of the correspondences below ${10^{-4}}$. P$_{\tt{proj}}$ indicates the reprojection error of correspondences less than $5$ pixels. Furthermore, considering the accuracy of pose estimation, we use the angular differences between the ground truth and predicted vectors for both translation and rotation as the error metric (AUC). Note that the AUC evaluation metric adopts the approximate AUC as in~\cite{zhang2019learning},~\cite{sun2021loftr} and~\cite{sgmnet}. Pose estimation is computed by first estimating the essential matrix with ${\tt findEssentialMat}$ (the threshold is set to $0.001$) implemented by OpenCV and RANSAC, followed by ${\tt recoverPose}$.

\subsection{Toy-Example Experiment}
\begin{figure}[!t]
\centering
   \includegraphics[width=1.0\linewidth]{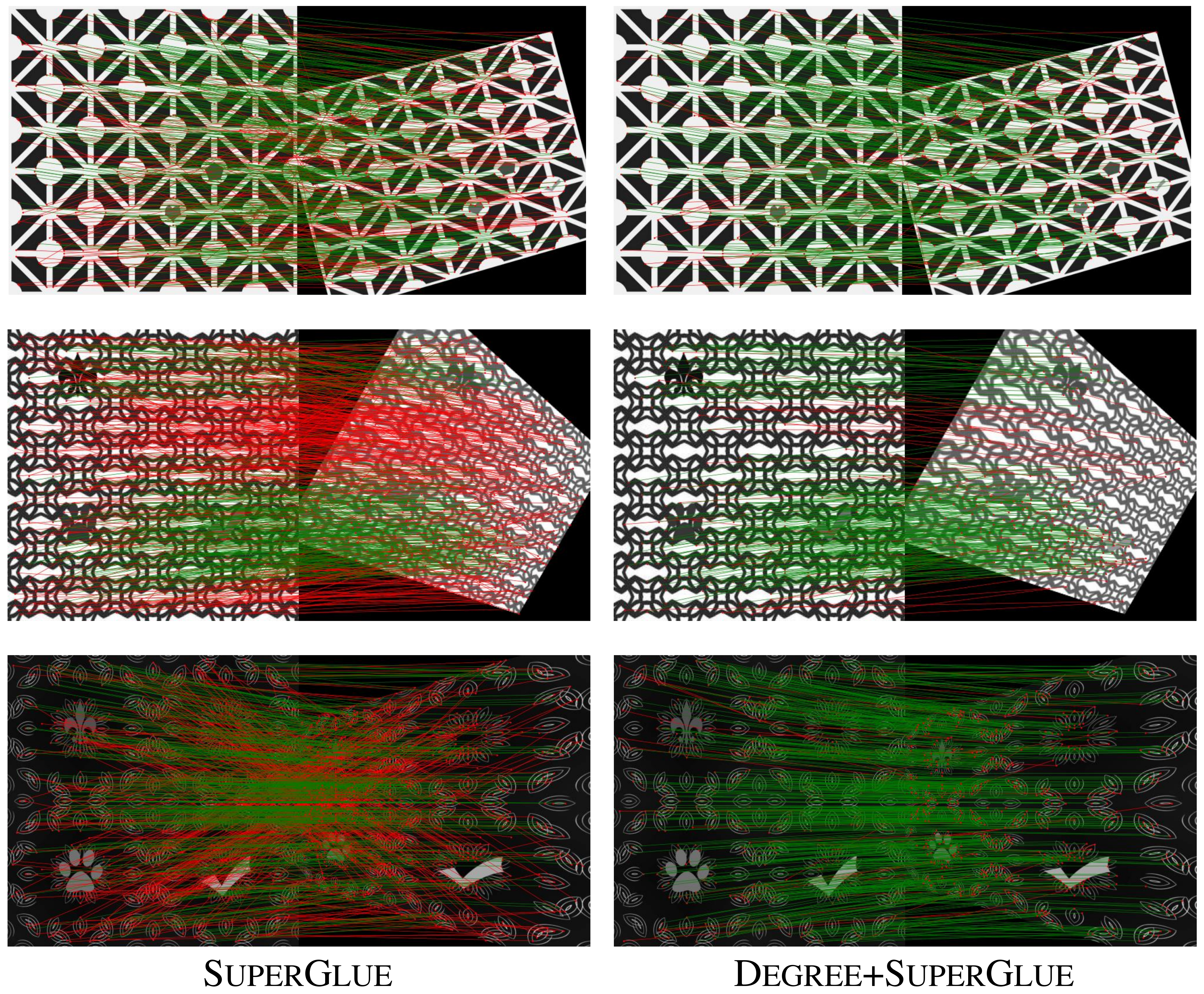}
   \caption{{Qualitative matching results with SuperPoint.} The first column is the result from SuperGlue, and the second column is the result from \ourmethod + SuperGlue. Correct matches are in green and mismatches are in red.}
\label{fig:toy experiment}
\end{figure}

To explore the performance of \ourmethod in the context of repeated patterns, we design a toy experiment. First, we select images that only contain one repeated structure, respectively. Then, three pattern marks are plotted on each image as anchor points for repeated patterns. For every image, a random homography matrix produces the corresponding image. The settings of the homography matrix are as follows:
The number of scales sampled during scaling is set to $5$. The number of angles sampled during rotation is set to $25$. The amplitude of scaling is $0.1$. The perspective effects in the $x$ direction and in the $y$ direction are $0.22$ and $0.25$, respectively. The proportion of the patch used to create the homography is $0.8$. The maximum rotation angle is $\pi$. The number of border artifacts due to translation is $1.0$. As shown in Fig.~\ref{fig:toy experiment}, \ourmethod shows a strong capability to correct mismatches with significant position errors. This controlled experiment highlights that geometric-invariant coordinates are vital for repeated patterns.

\begin{table*}[!t]
\centering
\renewcommand{\arraystretch}{1.2}
\addtolength{\tabcolsep}{0.5pt}
\caption{Evaluation on the HPatches dataset. We report the precision (P), recall (R), and percentage of correctly estimated homographies (Acc.) whose average corner error distance is below $4$ pixels.`Re-implemented' indicates the results obtained from our reimplementation using the official code; `Reported' indicates the results cited from the original paper.} 
\begin{tabular}{@{}lllcccccc@{}} 
\toprule
 \multirow{3}{*}{Descriptor} & \multirow{3}{*}{Method} & \multirow{3}{*}{Remark} & \multicolumn{3}{c}{Viewpoint} & \multicolumn{3}{c}{Illumination} \\ \cmidrule(lr){4-6} \cmidrule(lr){7-9}
     ~ &  ~  &  ~  &\makecell*[c]{P($\%$)}  & R($\%$) & Acc. ($\epsilon <4px$)    &P($\%$)  & R($\%$) & Acc. ($\epsilon <4px$)  \\ \hline
 \multirow{6}{*}{SuperPoint}  & NN+OANet~\citep{zhang2019learning}  &Re-implemented     &  89.5  &  82.4  &  67.0  &  \textbf{95.8}  &  83.6  & 91.7 \\
 ~  & SuperGlue~\citep{sarlin2020superglue}  &Reported &  91.4  &  \textbf{95.7}  &  --  &  89.1  &  91.7  & -- \\
 ~  & SuperGlue  &Re-implemented  &  90.4  &  92.8  &  73.7  &  91.6  &  93.3  & 93.7 \\
 ~ & SGMNet~\citep{sgmnet}  &Re-implemented  &  91.2  &  93.9  &  74.5  &  90.2  &  91.5  & 92.8 \\
 ~  & ClusterGNN~\citep{clustergnn} &Reported & --  &  -- &  74.0  &  --  &  --    & 93.0\\
 ~  & SuperGLue+\ourmethod    & --  &  \textbf{92.7}  &  93.2  &  \textbf{75.4}  &  91.1  &  \textbf{94.8}   & \textbf{94.6} \\
\bottomrule
\end{tabular}
\label{tab:hpatches}
\end{table*}

To further quantitatively evaluate \ourmethod, we performed an experiment on the HPathces dataset~\citep{balntas2017hpatches}. This dataset depicts planar scenes with ground-truth homographies and contains $56$ sequences with viewpoint changes and $52$ sequences with illumination changes. We follow the setup proposed in Patch2Pix~\citep{zhou2021patch2pix}, and report precision, recall, and the percentage of correctly estimated homography whose average corner error distance is below $3$ pixels. For all comparing methods, we apply the RANSAC implemented by OpenCV toolbox to estimate the homography matrix. For each image pair, we treat matches whose reprojection errors are less than $4$ pixels under the ground truth transformation as inliers. 
To make a fair comparison, we select SuperPoint~\citep{detone2018superpoint} as the local feature descriptor, choose $4$ K keypoints for all methods, and use the same hyper-parameters. Since ClusterGNN~\citep{clustergnn} does not provide the code, we directly cite their reported results. 
The HPatches experiments are summarized in Table~\ref{tab:hpatches}. Our method achieves competitive performance compared with other approaches~\citep{sarlin2020superglue,zhang2019learning,sgmnet,clustergnn}, especially with a significant improvement in homography estimation. This is attributed to the fact that each image pair in the Hpatches dataset contains only one single homography transformation, which is highly beneficial for our method.

\begin{figure}[!t]
\centering  
\subfigure[\ourmethod for SuperGlue]{
\includegraphics[width=0.48\textwidth]{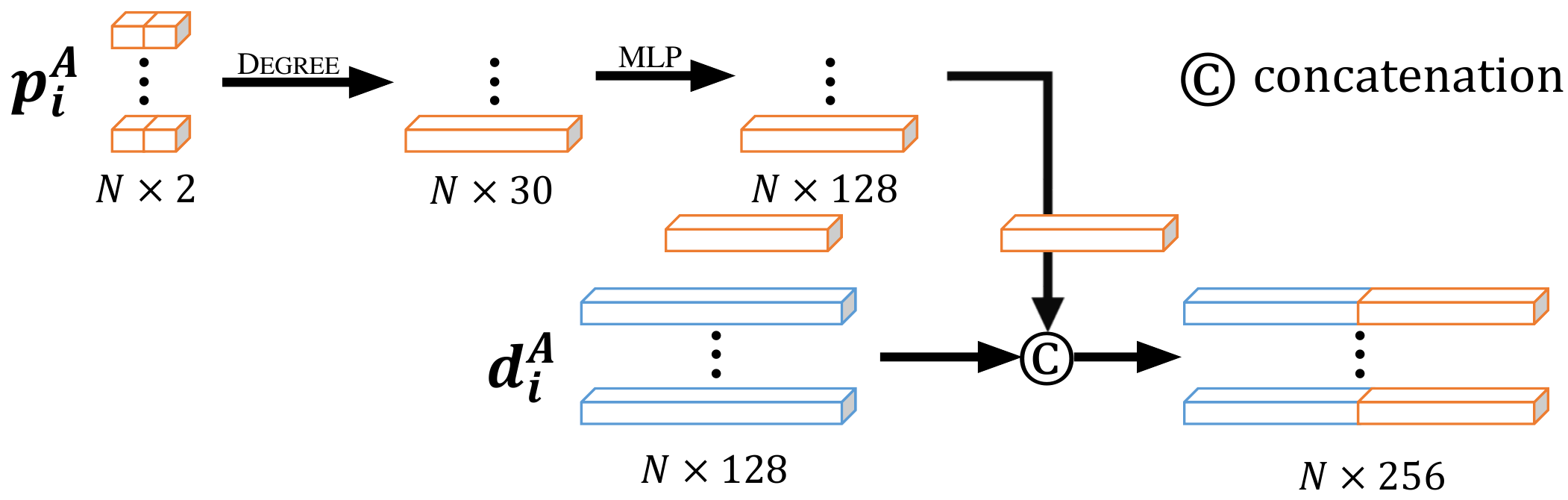}}
\quad
\subfigure[\ourmethod for OANet]{
\includegraphics[width=0.48\textwidth]{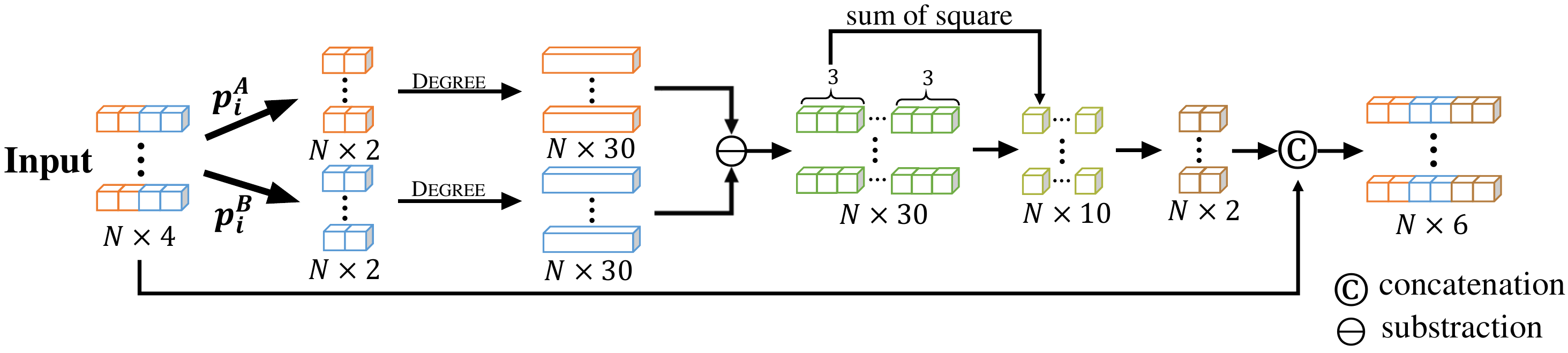}}
\caption{{Details of \ourmethod for different baselines.} (a) and (b) use five seed correspondences for \ourmethod.} 
\label{Fig:3}
\end{figure}


\subsection{Correspondence Coordinates for the Feature Matcher}

Recent SuperGlue is used as our feature matcher baseline. SuperGlue has a multilayer perceptron (MLP) implemented keypoint encoder with the layer setting of $(2, 32, 64, 128)$. To compare \mbox{\textsc{Degree}+SuperGlue} with SuperGlue, we choose the seed correspondences of $K=5$ to generate a coordinate of dimensions $30$ in Eq.~\eqref{eq:superglue} to replace the original coordinate representation. 
The coordinate is then embedded in a $128$ dimensional feature with also an MLP of $(30, 64, 128)$ for a fair comparison. Finally, we concatenate it with the input descriptor (Fig.~\ref{Fig:3}(a)). The rest of the network structure remains the same as that of SuperGlue. For the training setting, we follow~\cite{sarlin2020superglue} which uses Adam optimizer with a constant leaning rate of ${10^{-4}}$ for the first iterations $100k$, followed by an exponential decay of $0.999992$ until $600k$ iterations. We used a batch of $8$ image pairs.

\begin{figure}[!t]
\centering
   \includegraphics[width=1.0\linewidth]{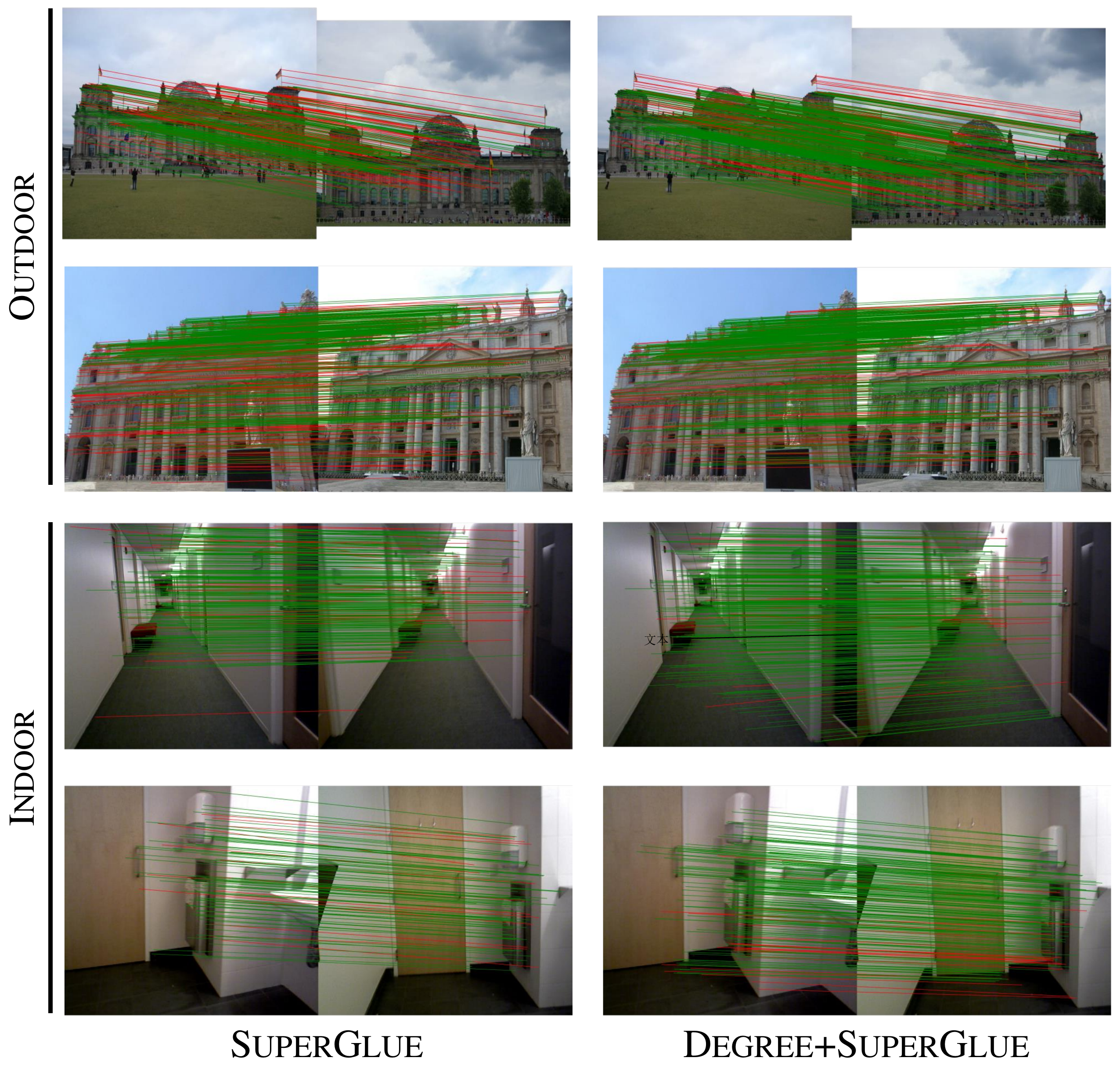}
   \caption{{Qualitative matching results with HardNet}.
   Compared with SuperGlue, \ourmethod+SuperGlue has more correct correspondences (green) and fewer mismatches (red). Best viewed in color and by zooming in.}
\label{fig:match vis}
\end{figure}
\begin{figure*}[!t]
\centering  
\includegraphics[width=0.95\linewidth]{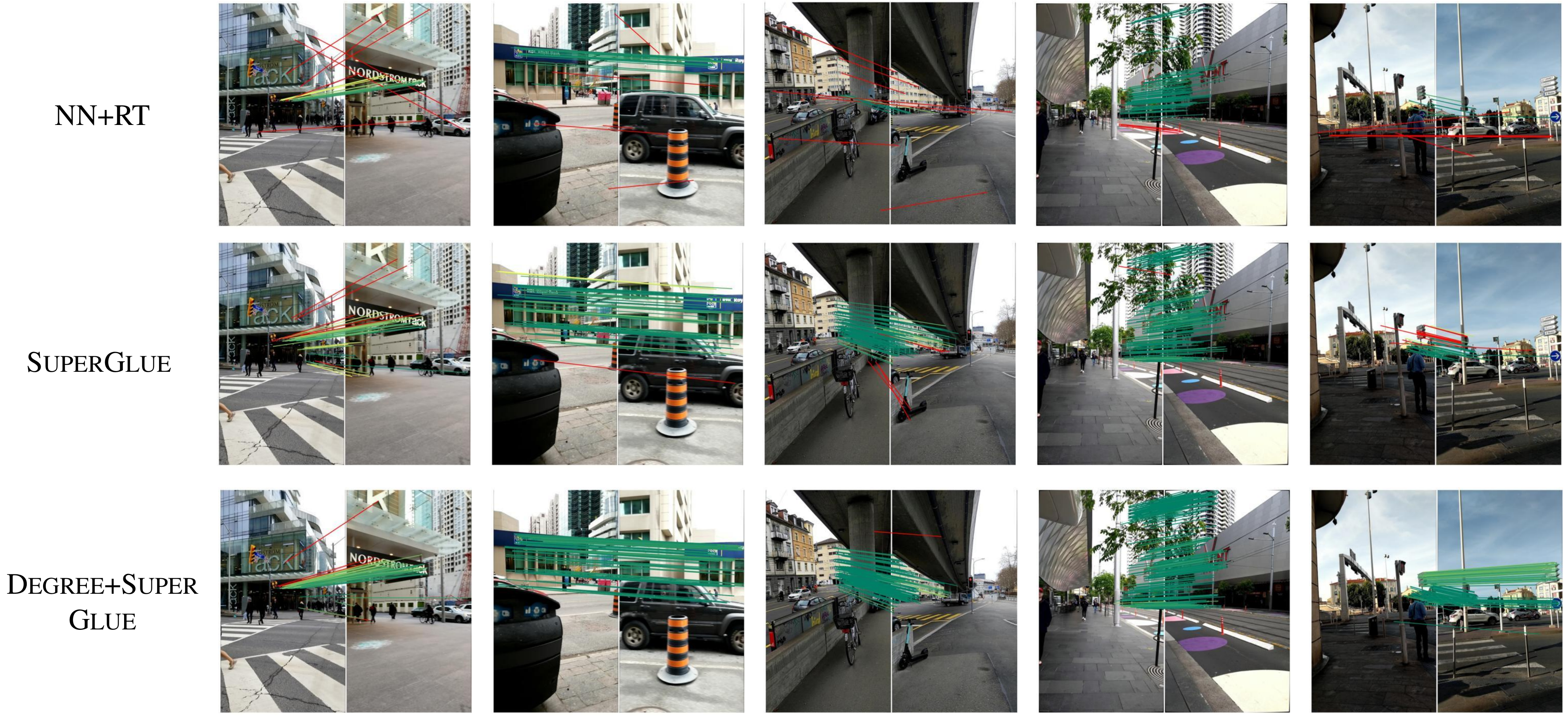}
\caption{Qualitative matching results with SuperPoint on GoogleUrban Dataset. It shows that our method performs better in scenes with low overlap.} 
\label{Fig:googleurban}
\end{figure*}
\begin{figure*}[!t]
\centering  
\includegraphics[width=0.95\textwidth]{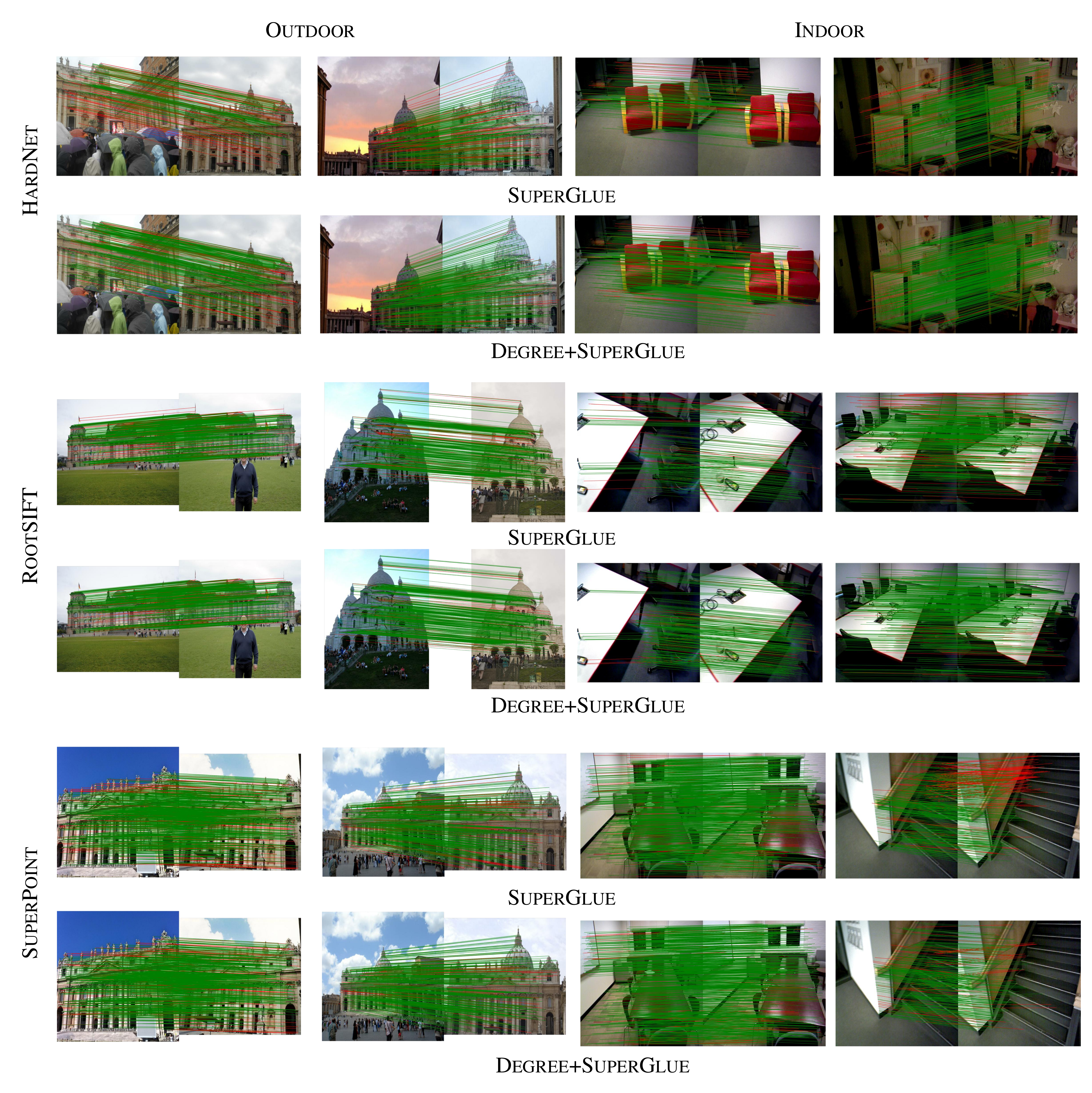}
\caption{{More visualization of various descriptors.} Correct matches are green lines and mismatches are red lines.} 
\label{Fig:4}
\end{figure*}

\vspace{5pt} \noindent\textbf{Outdoor Results.} 
Results are reported in Table~\ref{tab:superglue}. We observe that \mbox{\textsc{Degree}+SuperGlue} improves pose estimation and matching precision with different descriptors, especially when HardNet is used, as shown in Fig.~\ref{fig:match vis} (Additional visualizations can be found in Fig.~\ref{Fig:4}). Since HardNet does not encode any global information, SuperGlue can acquire only global information from the keypoint coordinates. However, SuperGlue with original Cartesian coordinates exhibits poor performance. When applying \ourmethod to SuperGlue, a considerable improvement is observed in all metrics ($+6.3\%$ in precision, $+11.0\%$ in recall and $+5.5\%$ in pose estimation), suggesting that it is the coordinate representation that matters. To better show the performance gap, we also visualize some results in Fig.~\ref{Fig:googleurban}. It shows that our method performs better in scenes with low overlaps, with more correct correspondences and fewer mismatches.
Note that the official training code of SuperGlue is not available. Its released model is trained with the SuperPoint descriptor. Despite carefully tuning the training strategy and inquiring about details from the authors, there still remains a performance gap between our reimplementation and the official reported results. Therefore, we inform the readers to compare the relative improvement.

\vspace{5pt} \noindent\textbf{Indoor Results.}
In indoor scenes, the improvement of \ourmethod is not as striking as in outdoor scenes. But \ourmethod still works in most cases. Similarly to outdoor cases, HardNet achieves the best results, with $+2.5\%$ in AUC@$20^{\circ}$ and $+4\%$ in recall. However, the SuperGlue with SuperPoint result does not perform as well as~\cite{sarlin2020superglue} achieved on ScanNet~\citep{dai2017scannet}. One reason is mentioned in~\cite{sarlin2020superglue} where camera poses in SUN3D are estimated from the sparse SIFT-based SfM, while ScanNet employs RGB-D fusion and optimization, providing significantly more accurate poses. Another reason may be the number of keypoints. We use $2$ K keypoints per image (to align with OANet), while~\cite{sarlin2020superglue} uses $400$ keypoints only. Indoor scenes often have few textures and require fewer keypoints. The large number of useless points may be the reason for the loss of accuracy in pose estimation.


\begin{table*}[!t]
\centering
\addtolength{\tabcolsep}{-2pt}
\renewcommand{\arraystretch}{1.2}
\caption{ Comparison with SuperGlue on the PhotoTourism and SUN3D datasets. P$_{\tt{proj}}$ denotes the mean matching precision with the projection distance error, and P$_{\tt{epi}}$ denotes the mean matching precision with the epipolar geometric distance error. `Released' indicates the results obtained from the officially released models; `Re-trained' indicates the models re-trained on the MegaDepth dataset~\citep{li2018megadepth}; `Re-implemented' indicates the results obtained from our re-implementation using the official code. Best performance is in bold. }
\begin{tabular}{@{}llllccccccc@{}}
\toprule
 &
   &
   &
   &
  \multicolumn{3}{c}{AUC} &
   &
   &
   &
   \\
\multirow{-2}{*}{Dataset} &
  \multirow{-2}{*}{Descriptor} &
  \multirow{-2}{*}{Matcher} &
  \multirow{-2}{*}{Remark} &
  @${5^{\circ}}$ &
  @${10^{\circ}}$ &
  @${20^{\circ}}$ &
  \multirow{-2}{*}{{P}$_{\tt{proj}}$ $(\%)$} &
  \multirow{-2}{*}{R $(\%)$} &
  \multirow{-2}{*}{F $(\%)$} &
  \multirow{-2}{*}{P$_{\tt{epi}}$ $(\%)$} \\ \midrule
 &
   &
  SuperGlue &
  Re-implemented &
  59.49 &
  67.67 &
  76.58 &
  71.27 &
  46.80 &
  55.67 &
  91.98 \\ 
 &
  \multirow{-2}{*}{RootSIFT} &
  \ourmethod+SuperGlue &
  -- &
  \textbf{61.18} &
  \textbf{69.28} &
  \textbf{77.98} &
  \textbf{74.43} &
  \textbf{52.65} &
  \textbf{60.87} &
  \textbf{94.87} \\ \cmidrule(l){5-11} 
 &
   &
  SuperGlue &
  Re-implemented &
  32.40 &
  42.21 &
  53.45 &
  55.81 &
  23.90 &
  31.74 &
  87.51 \\
 &
  \multirow{-2}{*}{HardNet} &
  \ourmethod+SuperGlue &
  -- &
  \textbf{36.65} &
  \textbf{47.06} &
  \textbf{58.94} &
  \textbf{62.15} &
  \textbf{34.89} &
  \textbf{42.98} &
  \textbf{92.36} \\ \cmidrule(l){5-11} 
  
  &
   &
  SuperGlue  &
  Released &
  63.42 &
  71.66 &
  79.33 &
  79.92 &
  64.05 &
  72.23 &
  95.94 \\ 
 &
   &
  SuperGlue &
  Re-implemented &
  61.23 &
  69.80 &
  78.79 &
  76.27 &
  63.89 &
  68.86 &
  {94.88} \\ 
\multirow{-10}{*}{PhotoTour.} &
  \multirow{-2}{*}{SuperPoint} &
  \ourmethod+SuperGlue &
  -- &
  {61.50} &
  {70.59} &
  {79.60} &
  {76.83} &
  {65.86} &
  {70.49} &
  94.80 
  \\
    &
   &
  \ourmethod+SuperGlue
  &Re-trained
  &\textbf{64.56} 
  &\textbf{73.57}
  &\textbf{81.64}
  &\textbf{80.54}
  &\textbf{69.10}
  &\textbf{75.98}
  &\textbf{97.22}
   \\ \midrule
\multicolumn{1}{l}{} &
   &
  SuperGlue &
  Re-implemented &
  {\textbf{13.04}} &
  {20.37} &
  {30.88} &
  {45.52} &
  {\textbf{18.88}} &
  {\textbf{24.27}} &
  {69.29} \\ 
\multicolumn{1}{l}{} &
  \multirow{-2}{*}{RootSIFT} &
  \ourmethod+SuperGlue &
  -- &
  {12.89} &
  {\textbf{20.62}} &
  {\textbf{31.63}} &
  {\textbf{48.06}} &
  {17.42} &
  {23.02} &
  {\textbf{70.32}} \\ \cmidrule(l){5-11} 
\multicolumn{1}{l}{} &
   &
  SuperGlue &
  Re-implemented &
  {12.04} &
  {19.10} &
  {29.55} &
  {45.42} &
  {16.37} &
  {21.56} &
  {68.87} \\ 
\multicolumn{1}{l}{} &
  \multirow{-2}{*}{HardNet} &
  \ourmethod+SuperGlue &
  -- &
  {\textbf{13.19}} &
  {\textbf{20.98}} &
  {\textbf{32.04}} &
  {\textbf{46.58}} &
  {\textbf{20.33}} &
  {\textbf{25.63}} &
  {\textbf{71.21}} \\ \cmidrule(l){5-11} 

  &
   &
  SuperGlue
  &Released
  &11.42 
  &18.81
  &30.93
  &47.24 
  &28.34
  &35.31
  &70.82
   \\ 
     &
   &
  SuperGlue &
  Re-implemented &
  12.30 &
  {19.93} &
  {31.33} &
  {\textbf{47.70}} &
  {24.37} &
  {29.23} &
  {68.91} \\ 
\multirow{-8}{*}{SUN3D} &
  \multirow{-3}{*}{SuperPoint} &
  \ourmethod+SuperGlue &
  -- &
  {\textbf{13.21}} &
  {\textbf{21.15}} &
  {\textbf{32.37}} &
  {47.21} &
  {\textbf{26.55}} &
  {\textbf{31.40}} &
  {\textbf{72.08}} \\ 
  \bottomrule
\end{tabular}

\label{tab:superglue}
\end{table*}

\vspace{5pt} \noindent\textbf{Wide-Baseline Results.} To further evaluate our method on the wide-baseline setting, the indoor pose estimation experiment is conducted on the ScanNet dataset~\citep{dai2017scannet}. Specifically, we split the testing image pairs into two clusters based on overlap ratios, including the easy cluster ($0.4 \sim 0.8$) and the hard one ($0.1 \sim 0.4$). We detect up to $2$ K SIFT keypoints (using the OpenCV implementation). We used the pre-trained model on the SUN3D dataset of SMFNet to extract seed correspondences. The comparison results are reported in Table~\ref{tab:scannet}. For different baselines, our method shows mostly competitive performance against state-of-the-art. Especially at low overlap ratios, our method maintains high matching precision ($+2.89\%$ compared to SuperGlue), which contributes to the final pose accuracy. Instead, SGMNet shows poor performance on the hard testing cluster.

\begin{table*}[!t] 
\centering
\renewcommand{\arraystretch}{1.2}
\addtolength{\tabcolsep}{1pt}
\caption{Wide-baseline indoor pose estimation on the ScanNet dataset~\citep{dai2017scannet}. `Re-implemented' indicates the results obtained from our reimplementation using the official code; `Released' indicates the results of the officially released model. }

\begin{tabular}{@{}llcccccc@{}} 
\toprule
 \multirow{3}{*}{Method} & \multirow{3}{*}{Remark} & \multicolumn{3}{c}{Easy} & \multicolumn{3}{c}{Hard}\\ \cmidrule(lr){3-5} \cmidrule(lr){6-8} 
  ~  &  ~ &\makecell*[c] AUC@$20^{\circ}$ & P($\%$)  & R($\%$)  & AUC@$20^{\circ}$ & P($\%$)  & R($\%$)  \\ \hline
 NN+RT  & Re-implemented    & 33.16   &  41.91  &  22.51  &  20.42  &  26.70  & 18.20 \\
 SuperGlue & Released  & 46.52   &  70.31  & \textbf{37.90}  &  27.24  &  43.36  & 25.83 \\
 SuperGlue & Re-implemented  & 45.41   &  69.13  &  35.78  &  25.84  &  41.53  & 25.29 \\
 SGMNet   & Re-implemented   & 46.03   &  70.48  & 37.40  &  23.56  &  38.09  & 24.17 \\
 \ourmethod+SuperGlue & --  & \textbf{46.71}   & \textbf{71.55}  &  36.97  &  \textbf{28.31}  &  \textbf{44.42}  & \textbf{27.08} \\
\bottomrule
\end{tabular}
\label{tab:scannet}
\end{table*}

\vspace{5pt} \noindent\textbf{Comparison on Pose Estimation.}
To compare our approach with other sparse matching methods, we conduct a comparative experiment on the YFCC100M dataset~\citep{thomee2016yfcc100m}. SMFNet and SuperGlue are both re-trained on the MegaDepth dataset~\citep{li2018megadepth} following SGMNet and ClusterGNN. We select the training image pairs based on the overlap rate from the SfM co-visibility and resize the longest side of the image to $1600$. The evaluation is performed on the same image pairs as in SuperGlue and ClusterGNN using the exact AUC. We report the results in Table~\ref{tab:yfcc100m comparison}. 
While our re-implemented SuperGlue slightly falls behind the official version, our approach still outperforms other methods with RootSIFT and SuperPoint. Specifically, compared with SGMNet and ClusterGNN, the clear performance advantage indicates the superiority of \ourmethod and the importance of appropriate coordinate representation for accurate correspondences.

\begin{table}[!t]  \scriptsize
\centering
\renewcommand{\arraystretch}{1.2}
\addtolength{\tabcolsep}{0pt}
\caption{Pose estimation on the YFCC100M dataset~\citep{thomee2016yfcc100m}. We compare different matching methods with different keypoint extraction approaches. All re-implemented models are re-trained on the MegaDepth dataset~\citep{li2018megadepth}. Best performance is in \textbf{boldface}.} 
\begin{tabular}{@{}lllcccc@{}}
\toprule
 \multirow{3}{*}{Descriptor}  & \multirow{3}{*}{Matcher}  & \multirow{3}{*}{Remark} & \multicolumn{3}{c}{AUC}   \\ \cmidrule(lr){4-6}
  ~  &  ~ & ~ & @$5^{\circ}$ & @$10^{\circ}$ & @$20^{\circ}$ \\ \hline
\multirow{5}{*}{RootSIFT} & SuperGlue &Reported & 30.49 & 51.29 & 69.72 \\
                       ~  & SuperGlue &Re-implemented & 31.07  & 50.31  & 68.94  \\
                       ~  & SGMNet  &Re-implemented  & 30.05  & 48.76  & 68.33 \\ 
                       ~  & ClusterGNN &Reported & \textbf{32.82}  & 50.25  & 65.89 \\
                       ~  & Ours &-- &32.66  & \textbf{52.73} & \textbf{71.24} \\\hline

\multirow{5}{*}{SuperPoint} & SuperGlue &Released & 38.57 & \textbf{59.23} & 75.61 \\
                         ~  & SuperGlue  &Re-implemented & 37.18 & 57.89 & 74.30  \\ 
                         ~  & SGMNet &Re-implemented & 33.30  & 52.64  & 70.17 \\ 
                         ~  & ClusterGNN &Reported & 35.31  & 56.13  & 73.56 \\
                         ~  & Ours &-- & \textbf{39.25} & 58.80 & \textbf{76.04}  \\
\bottomrule
\end{tabular}
\label{tab:yfcc100m comparison}
\end{table}

\subsection{Coordinates for the Consistency Filter}

The input of OANet is $N \times 4$ putative correspondences, which are the concatenation of paired coordinates. 
To compare \mbox{\textsc{Degree}+OANet} with OANet, we concatenate the two-dimensional feature vector in Eq.~\eqref{eq:oanet} with the original input (Fig.~\ref{Fig:3}(b)). All other settings are the same as for OANet. During training, we follow~\cite{zhang2019learning} to use the Adam solver with a learning rate of ${10^{-4}}$ and a batch size of $32$ pairs of images. 

The results are shown in Table~\ref{tab:oanet}. In both datasets, \ourmethod shows a performance improvement with all descriptors, especially for RootSIFT on YFCC100M. It achieves a $13\%$ improvement in precision and an $7.4\%$ improvement in recall. However, we observe an interesting phenomenon between RootSIFT and SuperPoint on YFCC100M: SuperPoint achieves better precision and recall than RootSIFT, while RootSIFT shows greater accuracy in pose estimation. A plausible explanation is that SuperPoint has good features but inaccurate keypoints.

To further demonstrate the efficacy of \ourmethod, we compare the proposed method with LAM~\citep{li2019lam}, an optimization-based outlier rejection method, on the Oxford dataset~\citep{oxford_dataset}. The Oxford dataset contains eight categories, including Bikes, Bark, Leuven, UBC, Trees, Boat, Graf, and Wall. Following the experimental settings in LAM, we use the affine-SIFT (ASIFT)~\citep{morel2009asift} algorithm implemented by VLFeat~\citep{vedaldi2010vlfeat} to generate initial correspondences, where the nearest-neighbor distance ratio (NNDR) is set to $0.85$. For each image pair, we treat matches whose reprojection errors are smaller than $3$ pixels under the ground truth transformation as inliers. We choose two baselines on the Oxford dataset -- OANet and OANet+\ourmethod, and additionally, report the mean absolute error (MAE) and root-mean-square error (RMSE) metrics. The results are shown in Table~\ref{tab:lam}. Compared with LAM, OANet+\ourmethod achieves a $2.96\%$ improvement in precision, an $3.97\%$ improvement in recall, a $0.14$ pixels improvement in MAE, and a $0.22$ pixels improvement in RMSE.

\begin{table*}[t]
\centering
\addtolength{\tabcolsep}{1pt}
\renewcommand{\arraystretch}{1.2}
\caption{Comparison with OANet on the YFCC100M and SUN3D datasets. NN denotes nearest-neighbor matching. The best performance is in boldface.}
\begin{tabular}{@{}lllcccccc@{}}
\toprule
 &
   &
   &
  \multicolumn{3}{c}{AUC} &
   &
   &
   \\ 
\multirow{-2}{*}{Dataset} &
  \multirow{-2}{*}{Descriptor} &
  \multirow{-2}{*}{Matcher} &
  @${5^{\circ}}$ &
  @${10^{\circ}}$ &
  @${20^{\circ}}$ &
  \multirow{-2}{*}{{P}$_{\tt{epi}}$ $(\%)$} &
  \multirow{-2}{*}{R $(\%)$} &
  \multirow{-2}{*}{F $(\%)$} \\ \midrule
 &
   &
  NN+OANet &
  54.60 &
  64.57 &
  74.25 &
  46.10 &
  79.50 &
  55.20 \\ 
 &
  \multirow{-2}{*}{RootSIFT} &
  \ourmethod+OANet &
  \textbf{55.47} &
  \textbf{64.79} &
  \textbf{76.55} &
  \textbf{59.30} &
  \textbf{87.15} &
  \textbf{68.42} \\ \cmidrule(l){4-9} 
 &
   &
  NN+OANet &
  56.20 &
  66.72 &
  76.91 &
  67.83 &
  87.85 &
  74.90 \\ 
 &
  \multirow{-2}{*}{HardNet} &
  \ourmethod+OANet &
  \textbf{57.21} &
  \textbf{67.84} &
  \textbf{78.13} &
  \textbf{69.23} &
  \textbf{88.33} &
  \textbf{76.08} \\ \cmidrule(l){4-9} 
 &
   &
  NN+OANet &
  41.83 &
  54.25 &
  \textbf{67.93} &
  64.63 &
  87.02 &
  72.04 \\ 
\multirow{-7}{*}{YFCC100M} &
  \multirow{-2}{*}{SuperPoint} &
  \ourmethod+OANet &
  \textbf{42.23} &
  \textbf{54.41} &
  67.63 &
  \textbf{65.01} &
  \textbf{87.35} &
  \textbf{72.47} \\ \midrule
 &
   &
  NN+OANet &
  {17.42} &
  {26.82} &
  {39.29} &
  {47.87} &
  {\textbf{84.59}} &
  {57.48} \\ 
 &
  \multirow{-2}{*}{RootSIFT} &
  \ourmethod+OANet &
  {\textbf{17.98}} &
  {\textbf{27.56}} &
  {\textbf{40.33}} &
  {\textbf{50.44}} &
  {84.17} &
  {\textbf{59.52}} \\ \cmidrule(l){4-9} 
 &
   &
  NN+OANet &
  {17.23} &
  {26.72} &
  {39.32} &
  {56.17} &
  {  86.57} &
  {  65.35} \\ 
 &
  \multirow{-2}{*}{HardNet} &
  \ourmethod+OANet &
  {\textbf{18.84}} &
  {\textbf{27.49}} &
  {\textbf{40.03}} &
  {\textbf{56.66}} &
  {\textbf{87.84}} &
  {\textbf{65.82}} \\ \cmidrule(l){4-9} 
 &
   &
  NN+OANet &
  {19.03} &
  {29.32} &
  {43.01} &
  {64.48} &
  {86.35} &
  {71.26} \\ 
\multirow{-7}{*}{SUN3D} &
  \multirow{-2}{*}{SuperPoint} &
  \ourmethod+OANet &
  {\textbf{19.88}} &
  {\textbf{30.86}} &
  {\textbf{44.18}} &
  {\textbf{65.13}} &
  {\textbf{87.26}} &
  {\textbf{73.89}} \\ \bottomrule
\end{tabular}
\label{tab:oanet}
\end{table*}

\begin{table}[!h] \scriptsize
\centering
\renewcommand{\arraystretch}{1.2}
\addtolength{\tabcolsep}{-1pt}
\caption{Performance comparison on the Oxford dataset. Best performance is in \textbf{boldface}} 

\begin{tabular}{@{}lccccc@{}}
\toprule
Method      & P$_{\tt{epi}}$ ($\%$) & R ($\%$) & F ($\%$) & MAE (pixels) & RMSE (pixels)  \\ \hline
LAM         & 92.84         & 95.61   & 93.75   & 1.48       & 1.74      \\
OANet       & 93.26         & 98.93   & 95.86   & 1.37       & 1.58     \\
OANet + \ourmethod  & \textbf{95.80}  & \textbf{99.58}   & \textbf{97.67}   & \textbf{1.34}       & \textbf{1.52}   \\ \bottomrule
\end{tabular}
\label{tab:lam}
\end{table}

\subsection{Ablation Study} \label{sec:ablation}

\vspace{5pt} \noindent\textbf{Reliability of Seed Correspondences.}
To generate reliable seed correspondences, we pre-train SMFNet on the PhotoTourism and SUN3D datasets with different descriptors. To alleviate the interference of repeated patterns, we apply a preprocessing to input keypoints. For the original input, we first apply Non-Maximum Suppression (NMS) with a radius of $4$ pixels, which discretizes the coordinate space distribution. Considering the extremely high similarity of the repeated patterns in terms of feature representation, a modified ratio test is implemented on the remaining descriptors, which further discretizes the feature space distribution. Concretely, we calculate the feature similarity of all descriptors within each image to obtain a similarity matrix, and then sort each row of the similarity matrix from large to small. If the $1$-st value in one of the rows is greater than $\gamma_1$ (the threshold for determining high similarity descriptors) and the ratio of the $6$-th value to the $1$-st value is greater than $\gamma_2$, we will discard this descriptor corresponding to this row. We apply different threshold settings to RootSIFT ($\gamma_1=0.94, \gamma_2=0.97$), HardNet ($\gamma_1=0.92, \gamma_2=0.93$), and SuperPoint ($\gamma_1=0.92, \gamma_2=0.88$).
We use Adam solver with a learning rate of ${10^{-4}}$ and a batch size of $8$ image pairs. The threshold $\alpha$ used for coarse matching is set to $0.1$. In $L_{\rm fine}$, $\lambda$ is set to $0$ during the first $20k$ iterations and then increases to $0.1$ in the rest of the $480k$ iterations.

\begin{figure}[!t]
    \centering
    \includegraphics[width=1.0\linewidth]{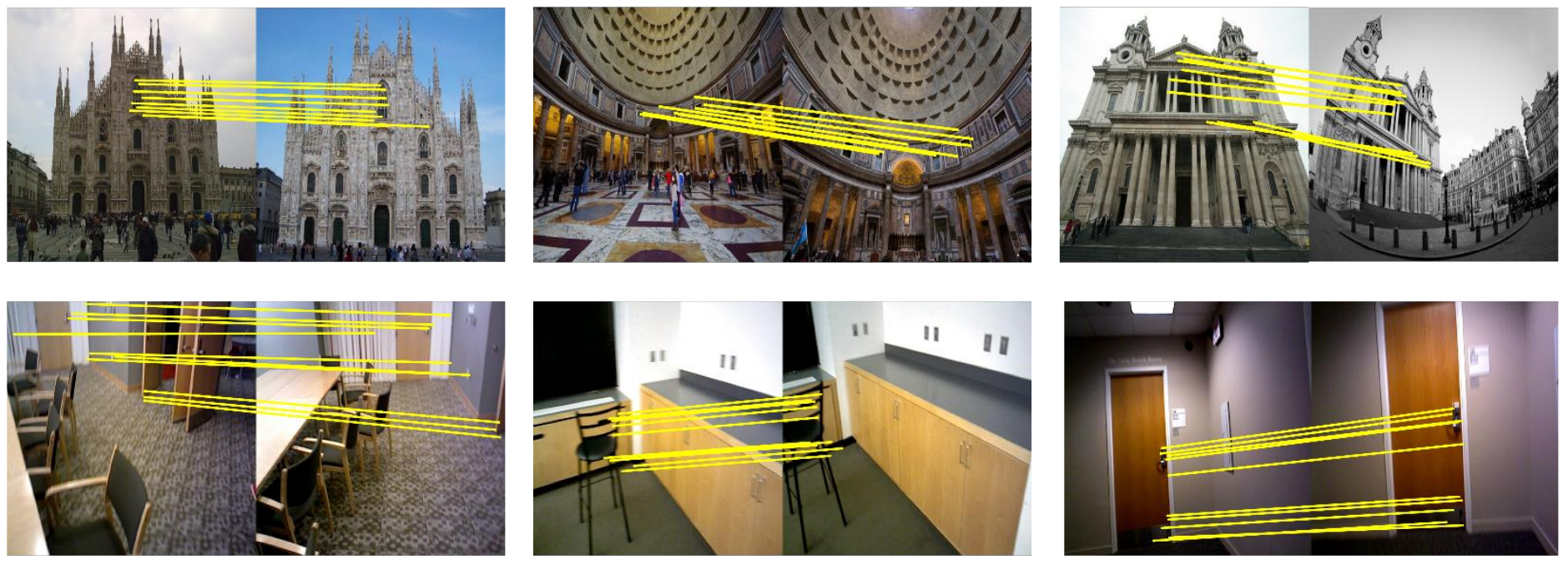}
    \caption{{Seed correspondences (anchors) predicted by our SMFNet.} Top-$10$ correspondences are visualized by yellow lines. We separately present the outdoor (top row) and indoor (last row) scenes.}
    \vspace{-10pt}
\label{fig:seed correspondences}
\end{figure}

The results in Table~\ref{tab:smfnet} show that pretraining can effectively help generate reliable seed correspondences, with at least $8$ correct correspondences in the top-$10$ confident candidates. 
To further demonstrate the benefits of SMFNet, we also compare it with simple NN matching followed by Lowe's ratio test~\citep{lowe2004distinctive}. Since the top $10$ matches are retained to evaluate the `Top-$10$' metric, we rank the correspondences obtained by the ratio test according to the feature similarity value from large to small. As shown in Table~\ref{tab:smfnet}, the `NN+RT' baseline has lower recall and precision. In particular, the `Top-10' metric points out the superiority of SMFNet in acquiring seed correspondences.
To visualize the quality of seed correspondences, we plot the top$10$ correspondences in Fig.~\ref{fig:seed correspondences}. We observe that correspondences mainly cluster around asymmetric/low-repetition areas in texture-rich outdoor scenes. In indoor scenes with few textures, correspondences mainly cluster around object edges. This is in line with our expectations.

\vspace{5pt} \noindent\textbf{Distribution of Seed Correspondences.}

\begin{figure*}[bhtp]
\makeatletter
\renewcommand{\@thesubfigure}{\hskip\subfiglabelskip}
\makeatother
\centering  
\subfigure{
\includegraphics[width=0.30\textwidth]{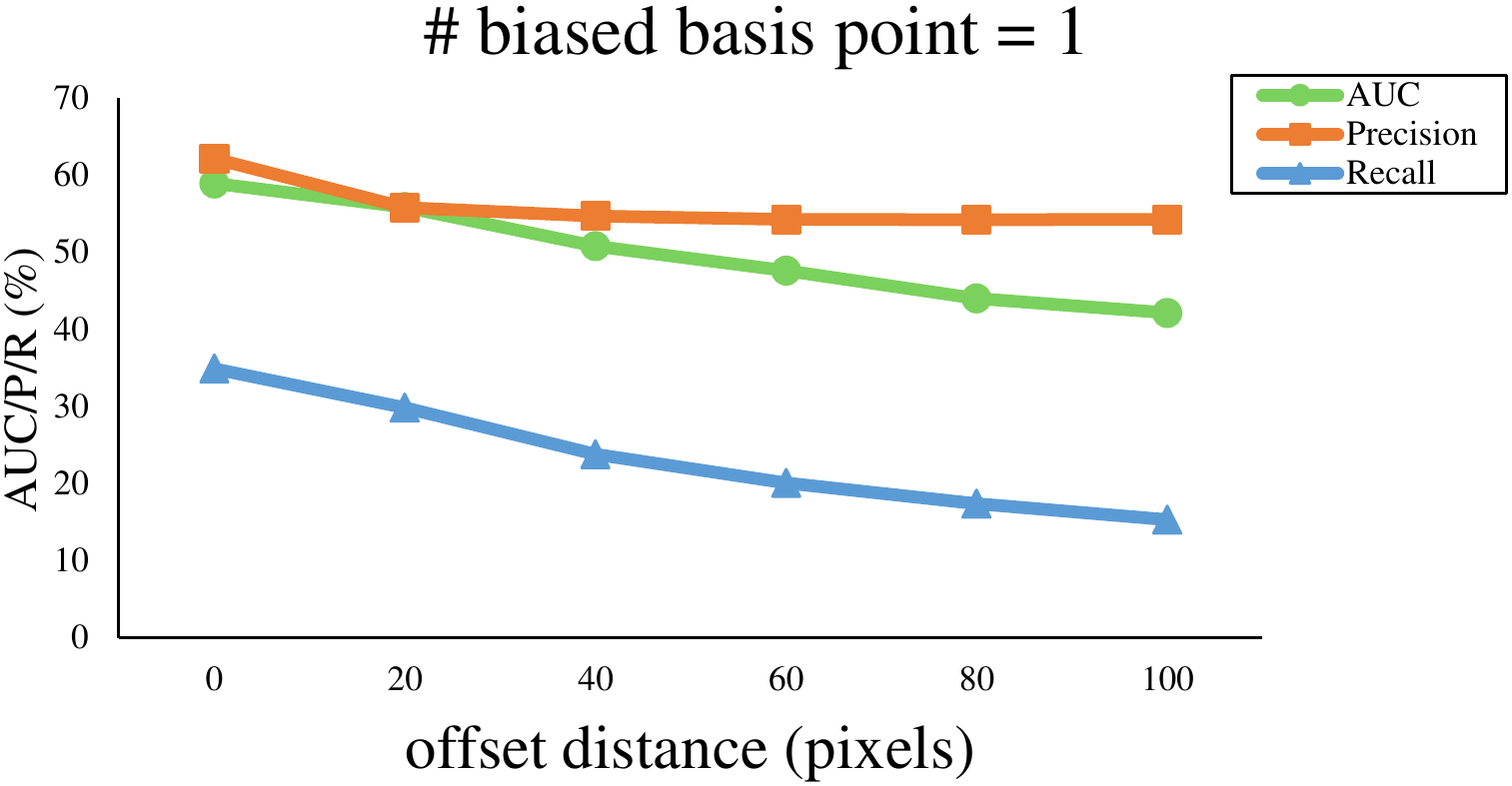}}
\quad
\subfigure{
\includegraphics[width=0.30\textwidth]{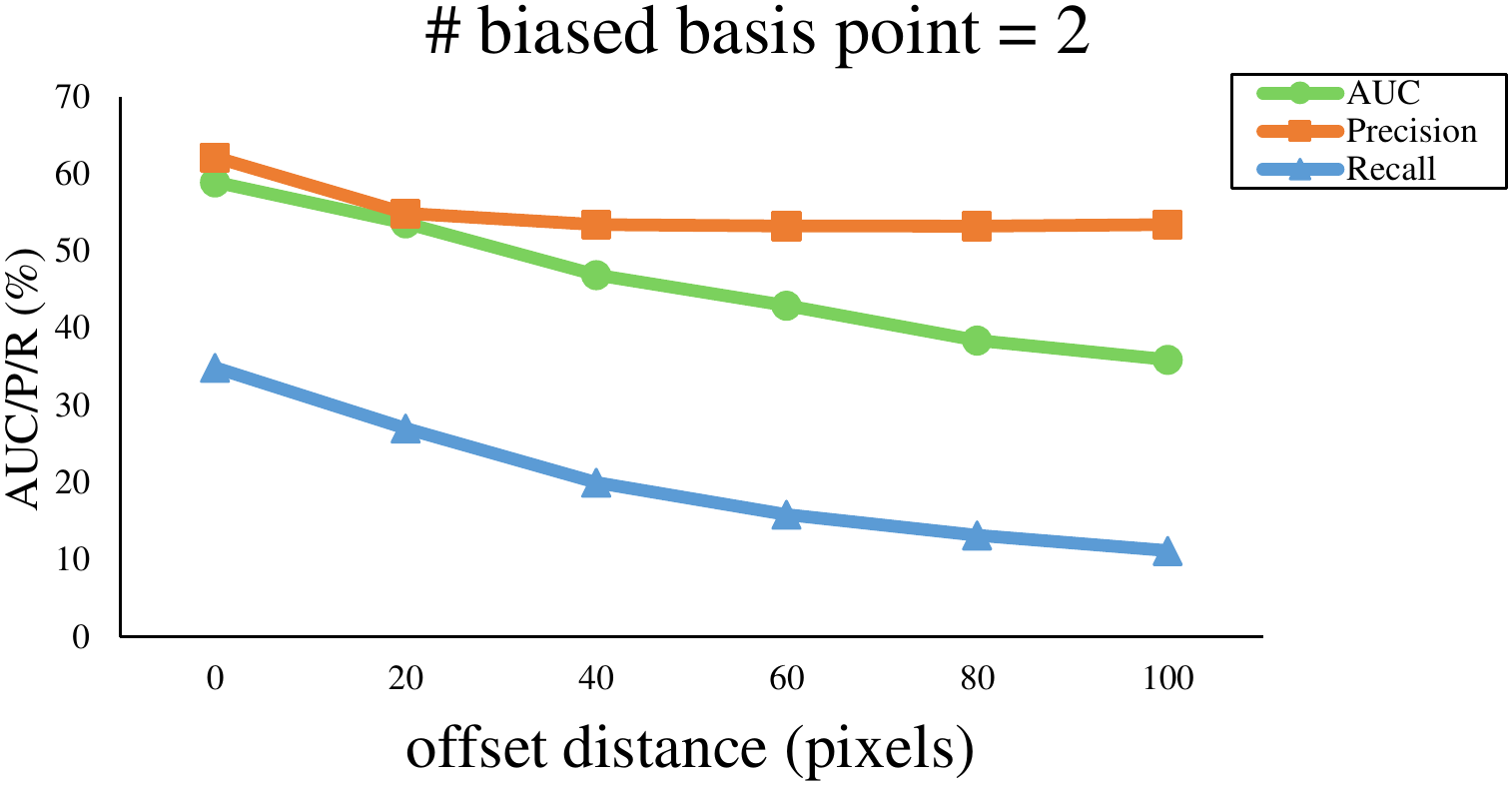}}
\quad
\subfigure{
\includegraphics[width=0.30\textwidth]{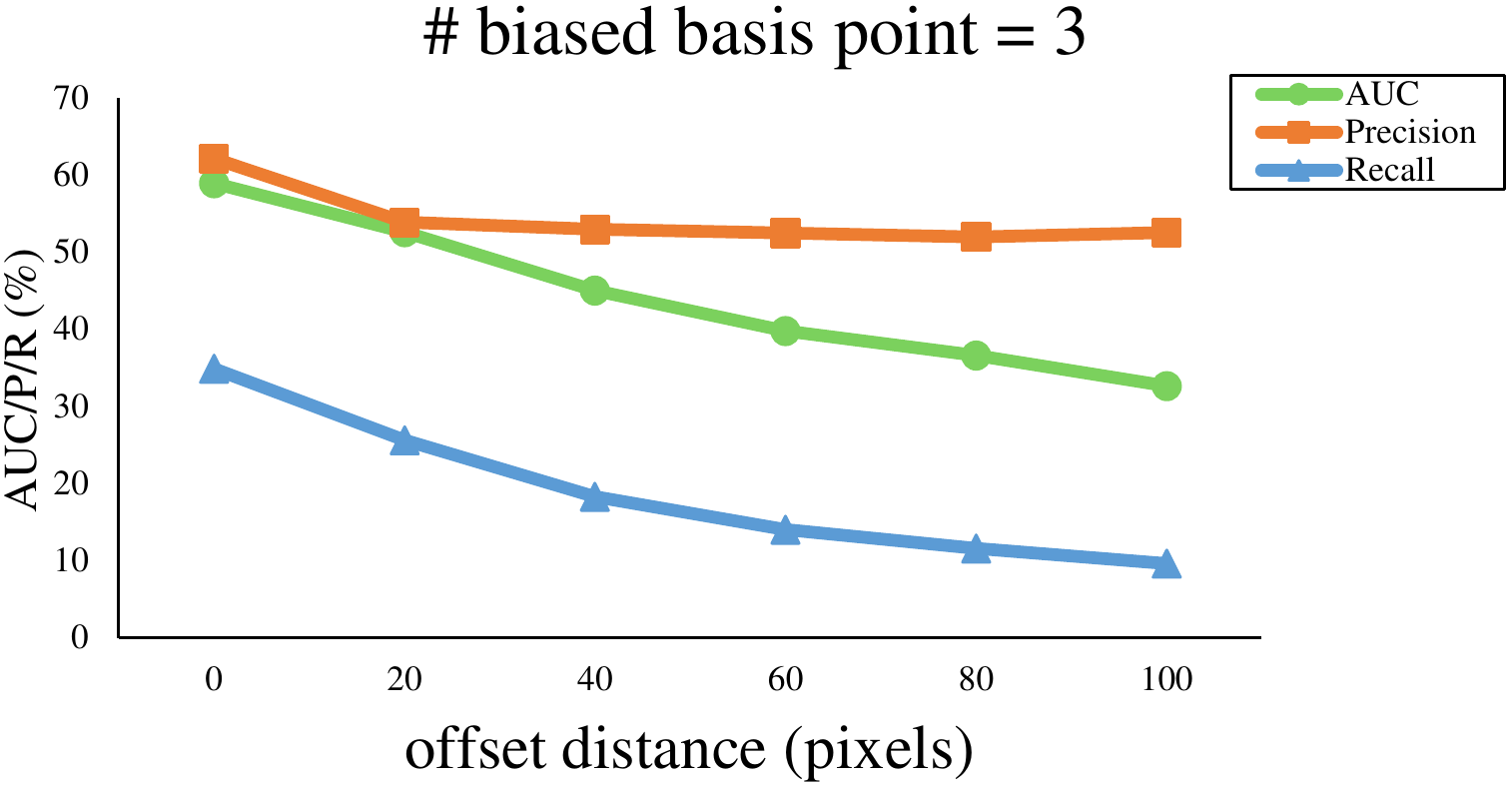}}
\caption{{Analysis of seed correspondence's accuracy.} We experiment with different numbers of mismatches and deviation distances using five seed correspondences. AUC represents AUC(@$20^{\circ}$). } 
\label{Fig:1}
\end{figure*}

\begin{figure*}[!t]
\centering  
\includegraphics[width=1.0\textwidth]{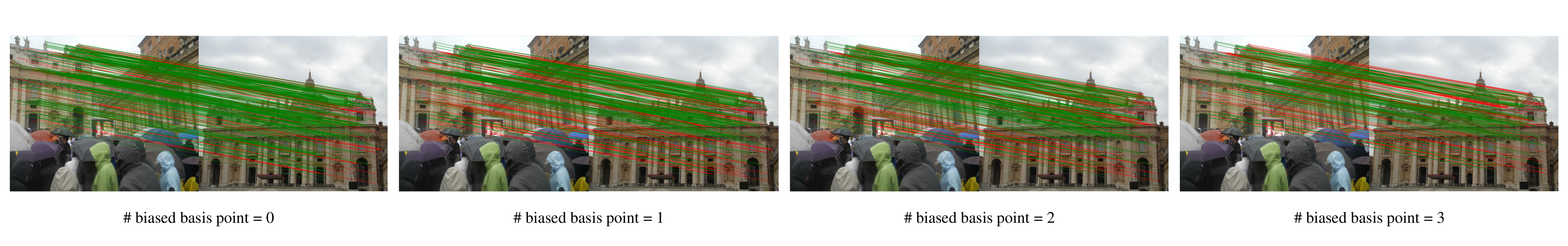}
\caption{{Visualization of different numbers of mismatch.} {{Green}} indicates correct matches, and {{red}} represents mismatches. } 
\label{Fig:2}
\end{figure*}

\begin{figure}[!t]
\centering  
\includegraphics[width=1.0\linewidth]{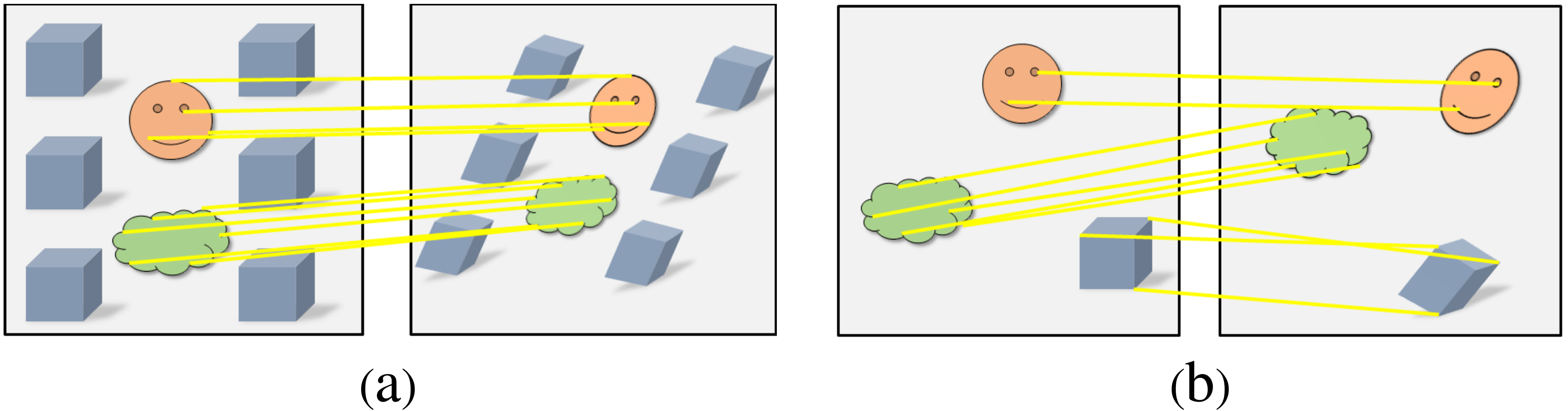}
\caption{Toy experiment of the distribution of seed correspondences.} 
\label{Fig:distribution}
\end{figure}

One may be interested in knowing the distribution of seed correspondences, because the distribution can affect the reliability of different barycentric coordinate systems. If multiple seed correspondences cluster in a local region, the benefit of the multiple coordinate systems weakens. Ideally, we expect a uniform distribution.
For a better and intuitive understanding of the seed distribution, we design another toy experiment to explain it. In Fig.~\ref{Fig:distribution}(a), due to the existence of many identical cubes in each image, the seed correspondences appear only on the single smiley-face shape and the single cloudy shape. In Fig.~\ref{Fig:distribution}(b), each shape appears only once and the correspondences are distributed uniformly. Additionally, the cloudy shape does not have symmetry and obvious repetitive patterns compared to the other two shapes. In this case, it can host more keypoints and seed correspondences.
Hence, the distribution of seed correspondences is highly data-dependent, but in most cases it should distribute uniformly if there exist multiple non-repetitive regions.

\begin{table}[!t]
\centering
\caption{Ablation of SMFNet. Both ACNet block and NMNet block are critical for higher accuracy, and a
refine network further improves the precision and recall.}
\begin{tabular}{@{}llccc@{}}
\toprule
Matcher                 & Setting                     & Top-10   &  {P}$_{\tt{epi}}$ $(\%)$  & R $(\%)$ \\ \midrule
NN &-- & 5.69 &0.671 & 0.632      \\ \hline
\multirow{4}{*}{SMFNet} & No NM-block      & 8.91           & 0.869      & 0.807  \\
                        & No AR-block     & 8.73          & 0.858      & 0.789  \\
                        & No Fine Matching          & 6.17           & 0.781      & -      \\ 
                        & Full & \textbf{9.88}  & \textbf{0.944} & \textbf{0.864}  \\
\bottomrule
\end{tabular}
\label{tab:smfnet ablation}
\end{table}

\begin{table}[!t] \scriptsize
\centering
\caption{Ablation study of position encoding in coarse-level matching. The positions are encoded by a series of MLP layers following~\cite{sarlin2020superglue} and~\cite{sgmnet}. }
\begin{tabular}{@{}llcccc@{}}
\toprule
Matcher                 & Setting   & Top-10 &{P}$_{\tt epi}$ $(\%)$  & R $(\%)$  & Time\\ \midrule
\multirow{2}{*}{SMFNet} & w. position encoding     & \textbf{9.91}  & \textbf{0.951} & 0.862 & +9ms  \\
                        & w.o. position encoding   & 9.88  & 0.944  &\textbf{0.864} & -- \\
\bottomrule
\end{tabular}
\label{tab:position encoder}
\end{table}

\begin{figure}[!t]
\centering  
\includegraphics[width=0.49\textwidth]{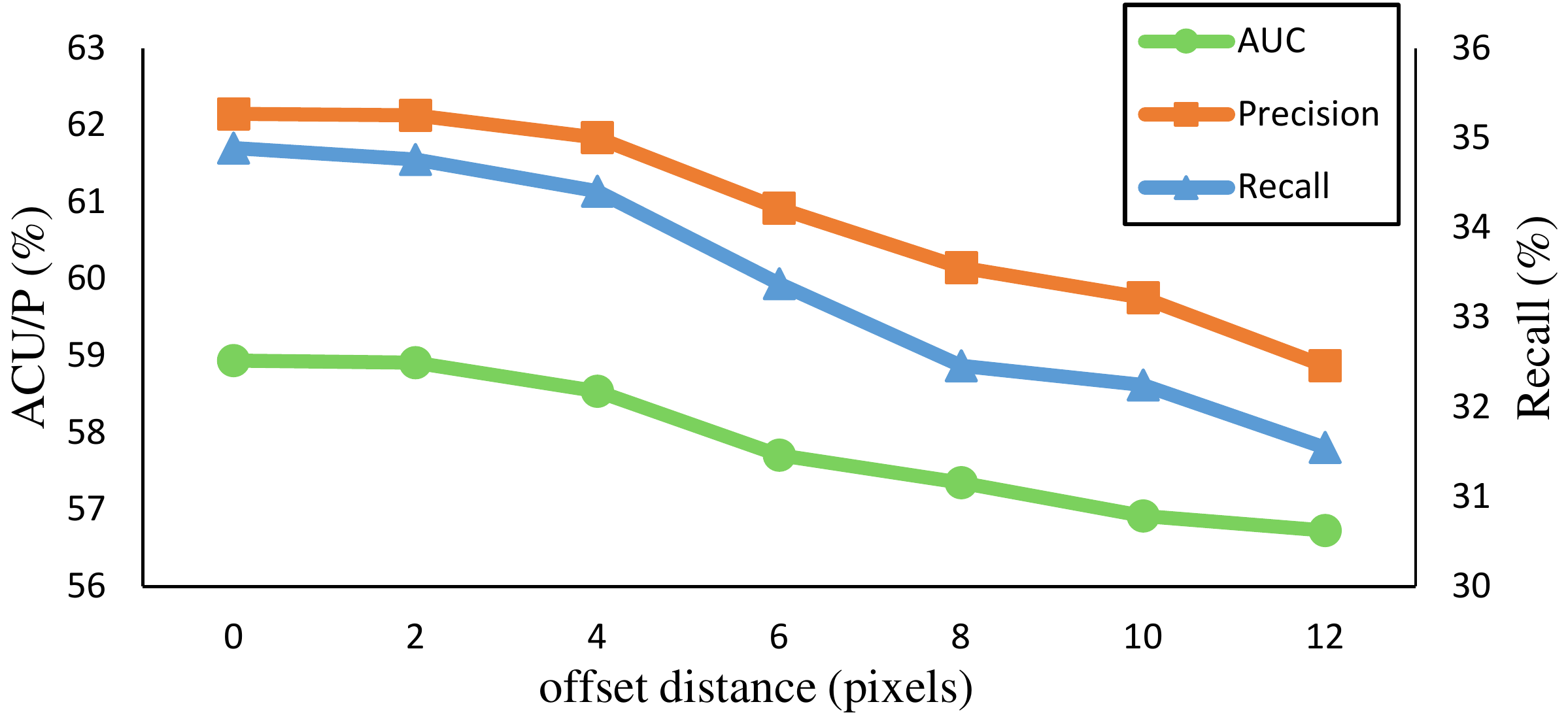}
\caption{{Analysis of the robustness of seed correspondences.} We experiment with biased basis point = $1$. We report AUC(@$20^{\circ}$), precision, and recall. } 
\label{Fig:analysis seed}
\end{figure}

\vspace{5pt} \noindent\textbf{Accuracy of Seed Correspondences.}
To further explore the impact of seed correspondence accuracy on matching performance, we manually add an offset with a fixed distance but a random direction to one basis point of a seed correspondence composing the coordinate system. SuperGlue with HardNet is used as a baseline, and the results are reported in the PhotoTourism dataset. As shown in Fig.~\ref{Fig:1}, we find that 
increased offsets lead to decreased precision, recall, and AUC. Additionally, as the number of biased basis points increases, the margins between precision and recall/AUC increase, implying that the accuracy of seed correspondences mainly affects recall. We visualize in Fig.~\ref{Fig:2} the correspondence results with different numbers of biased basis points w.r.t. the offset distance of $20$ pixels. It is clear that biased seed correspondences deteriorate matching performance. To further illustrate robustness to small errors, we change the offset distances ($0$/$2$/$4$/$6$/$8$/$10$/$12$ pixels) of one basis point. As shown in Fig.~\ref{Fig:analysis seed}, it can be observed that small errors ($\le 6$ pixels) have little impact on the matching performance.

\begin{figure}[!t]
    \centering
     \includegraphics[width=0.95\linewidth]{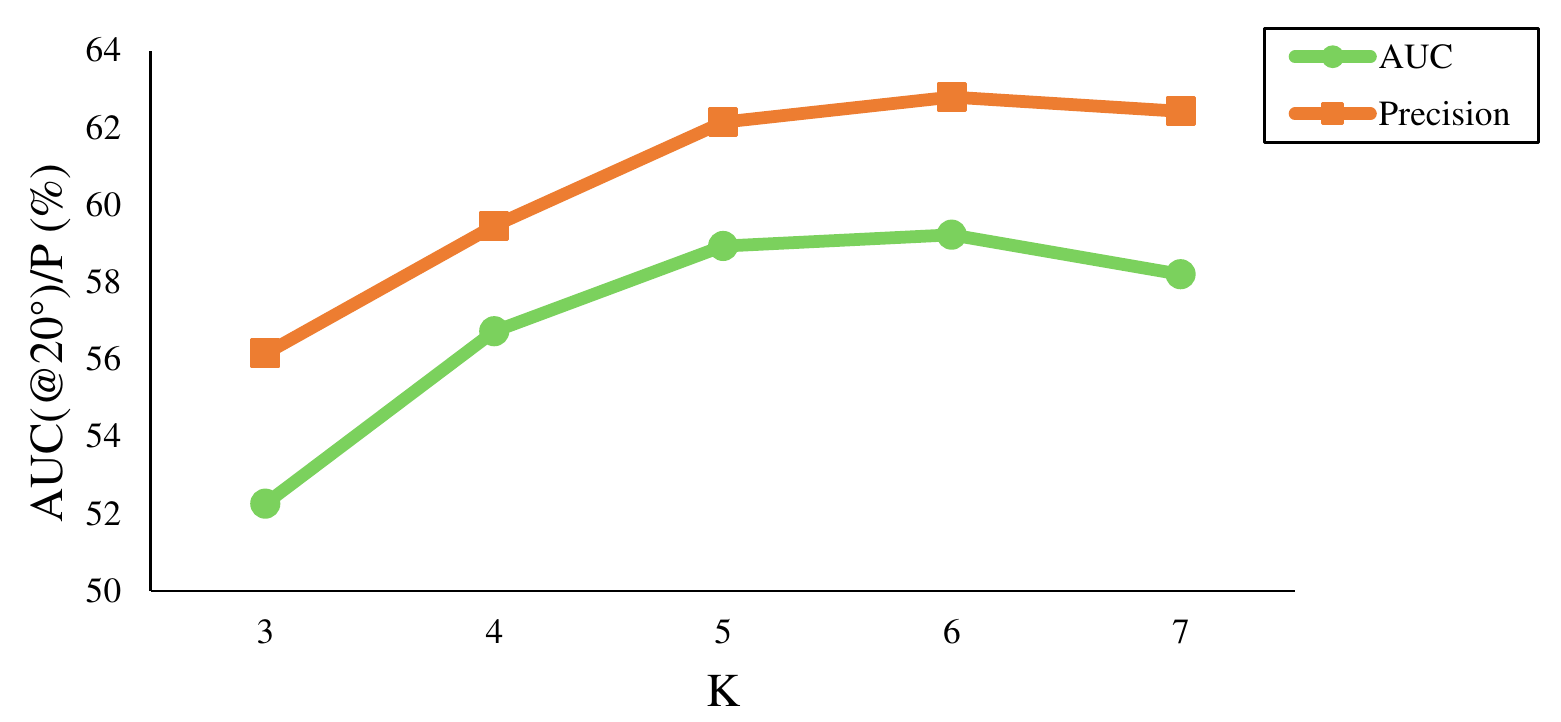}
    \caption{Ablation study on the number of seed correspondences ${K}$. We test \ourmethod+SuperGlue.
    }
\label{fig:ablation}
\end{figure}

\vspace{5pt} \noindent\textbf{The Number of Seed Correspondences.} 
To assess the impact of the number of seed correspondences ${K}$, we perform an ablation study on the YFCC100M dataset with SuperGlue and HardNet. As shown in Fig.~\ref{fig:ablation}, despite the fact that the best performance is achieved when ${K}=6$, ${K}=5$ has similar performance but with half the computational cost. Therefore, we recommend the choice of $K=5$.

\vspace{5pt} \noindent\textbf{SMFNet.} 
To justify our design choices in SMFNet, we repeat the outdoor experiments with HardNet features, but focus on different variants of SMFNet. Our results, presented in Table~\ref{tab:smfnet ablation}, show that all blocks are useful and bring substantial performance gains. 
To explain why we discard the position encoding during coarse matching in Sec.~\ref{subsec:coarse-level matchinag}, we give an explanation in Table~\ref{tab:position encoder}. Position encoding only brings a small performance improvement in precision but requires an additional $9$ ms of inference time. After weighing the costs and benefits, we do not apply position encoding to coarse matching.

\vspace{5pt} \noindent\textbf{Coordinate dimension of input keypoints.}
Considering that input key points use different coordinate dimensions between the compared methods and \ourmethod, here we conduct additional experiments to prove the fairness of \ourmethod. Compared to SuperGlue, the input coordinate dimension of \ourmethod is $30$. Thus, we lift the $2$-dimensional coordinates of SuperGlue to a $30$-dimensional vector using the standard positional encoding used in Transformer~\citep{vaswani2017attention,sun2021loftr}. The results are shown in Table~\ref{tab:superglue-lifted}. It shows that the lifted dimension even leads to a degradation in the AUC and precision. This phenomenon has also been pointed out by LoFTR~\citep{sun2021loftr}.
Similarly in OANet, the input coordinate dimension of \ourmethod is $6$. Thus, we lift the $4$-dimensional vector of OANet to a $6$-dimensional vector using the same strategy presented in Section~\ref{sec:4-3}. Specifically, we calculate the offset values of the coordinates of the two points in the correspondence and concatenate them behind the original input vector. The results are shown in Table~\ref{tab:oanet-lifted}.  Similar poor performance can be observed, suggesting that \textit{the way for encoding coordinates outweighs simply including it.}

\begin{table*}[!t]
\centering
\renewcommand{\arraystretch}{1.2}
\addtolength{\tabcolsep}{1pt}
\caption{Performance comparison with different coordinate representations. The best performance is in \textbf{boldface}.}
\begin{tabular}{@{}lllcccc@{}}
\toprule
Dataset    & Descriptor  & Matcher  &AUC@$20^{\circ}$ & P($\%$) & R($\%$) & F($\%$)  \\ \hline
\multirow{3}{*}{PhotoTour.}   &\multirow{3}{*}{RootSIFT} & SuperGlue  & 76.58  & 71.27  & 46.80  & 55.67 \\
~  &  ~  & SuperGlue + $30$D-sinusoidal embedding   &74.71   & 70.18   & 47.55   & 55.24  \\ 
~  &  ~  & SuperGlue + \ourmethod &\textbf{77.98} & \textbf{74.43} & \textbf{52.65} &\textbf{60.87} \\ \hline

\multirow{3}{*}{SUN3D}   &\multirow{3}{*}{RootSIFT} & SuperGlue  & 30.88 & 45.52 & \textbf{18.88} & \textbf{24.27} \\
~  &  ~  & SuperGlue + $30$D-sinusoidal embedding   & 28.30  & 43.45   &17.63    & 21.86   \\ 
~  &  ~  & SuperGlue + \ourmethod & \textbf{31.63}  &\textbf{48.06}  & 17.42  & 23.02  \\ 
\bottomrule
\end{tabular}
\label{tab:superglue-lifted}
\end{table*}

\begin{table*}[!t] 
\centering
\renewcommand{\arraystretch}{1.2}
\addtolength{\tabcolsep}{1pt}
\caption{Performance comparison of input vectors with different dimensions. Best performance is in \textbf{boldface}. OANet$^{\dag}$ denotes the modified version which lifts the $4$D vector to $6$D.} 
\begin{tabular}{@{}lllcccc@{}}
\toprule
Dataset    & Descriptor  & Matcher  &AUC@$20^{\circ}$ & P $_{\tt epi}$ ($\%$) & R ($\%$) & F ($\%$)  \\ \hline
\multirow{3}{*}{YFCC100M}   &\multirow{3}{*}{RootSIFT} & OANet ($4$D) & 74.25 & 46.10 & 79.50 & 55.20 \\
~  &  ~  & OANet$^{\dag}$ ($6$D)   & 55.70  & 39.61   & 67.46   & 48.83  \\ 
~  &  ~  & OANet + \ourmethod ($6$D) & \textbf{76.55} & \textbf{59.30} & \textbf{87.15} & \textbf{68.42} \\ \hline

\multirow{3}{*}{SUN3D}   &\multirow{3}{*}{RootSIFT} & OANet ($4$D) & 39.29 & 47.87 & \textbf{84.59} & 57.48 \\
~  &  ~  & OANet$^{\dag}$ ($6$D)  & 24.55  & 28.75   & 61.24   &  42.61  \\ 
~  &  ~  & OANet + \ourmethod ($6$D) & \textbf{40.33} & \textbf{50.44} & 84.17 & \textbf{59.52}  \\ 
\bottomrule
\end{tabular}
\label{tab:oanet-lifted}
\end{table*}

\vspace{-1pt}
\subsection{Pose Estimation with SMFNet}

To show that SMFNet is a sufficiently good choice for generating seed correspondences, here we present an application of SMFNet in pose estimation and also compare it with other approaches. 
Specifically, we use the correspondences obtained by SMFNet with confidence values greater than zero to calculate the camera pose. The results of the PhotoTourism dataset are shown in Table~\ref{tab:SMFNet:pose estimation}. SMFNet is on par with SuperGlue in the AUC metric, but outperforms it in precision. We think the reason is that SMFNet only generates sparse correspondences and cannot guarantee high recall. 

\begin{table}[!t] 
\centering
\caption{Pose estimation with SMFNet on the PhotoTourism dataset. We tested SMFNet and other methods with $2k$ keypoints. RT: ratio test. The best performance is in boldface. `Re-implemented' indicates the results obtained from our reimplementation using the official code; `Released' indicates the results of the officially released model.} 
\renewcommand{\arraystretch}{1.2}
\addtolength{\tabcolsep}{-1pt}
\begin{tabular}{@{}lllcc@{}}
\toprule
Descriptor                  & Matcher  & Remark  & AUC@${20^{\circ}}$ & P$_{\tt epi}$ $(\%)$              \\ \midrule
\multirow{3}{*}{RootSIFT}   & NN+RT     & Re-implemented  & 43.28          & 69.64          \\
                            & SuperGlue & Re-implemented  & \textbf{76.58} & 91.98          \\
                            & SMFNet    & --  & 74.53          & \textbf{96.77} \\ \hline
\multirow{3}{*}{HardNet}    & NN+RT     & Re-implemented  & 44.71         & 71.36  \\
                            & SuperGlue & Re-implemented  & 53.45          & 87.51          \\
                            & SMFNet    & --  & \textbf{53.66} & \textbf{90.87} \\ \hline
\multirow{4}{*}{SuperPoint} & NN+RT     & Re-implemented  & 40.90          &  65.34  \\
                            & SuperGlue & Released  & \textbf{79.33} & 95.94       \\
                            & SuperGlue & Re-implemented  & 78.79 & 94.88          \\
                            & SMFNet    & --  & 76.42          & \textbf{96.36} \\ 
\bottomrule
\end{tabular}
\label{tab:SMFNet:pose estimation}
\end{table}

\begin{figure}[!t]
    \centering
    \subfigure[Inference time]{
    \includegraphics[width=0.43\textwidth]{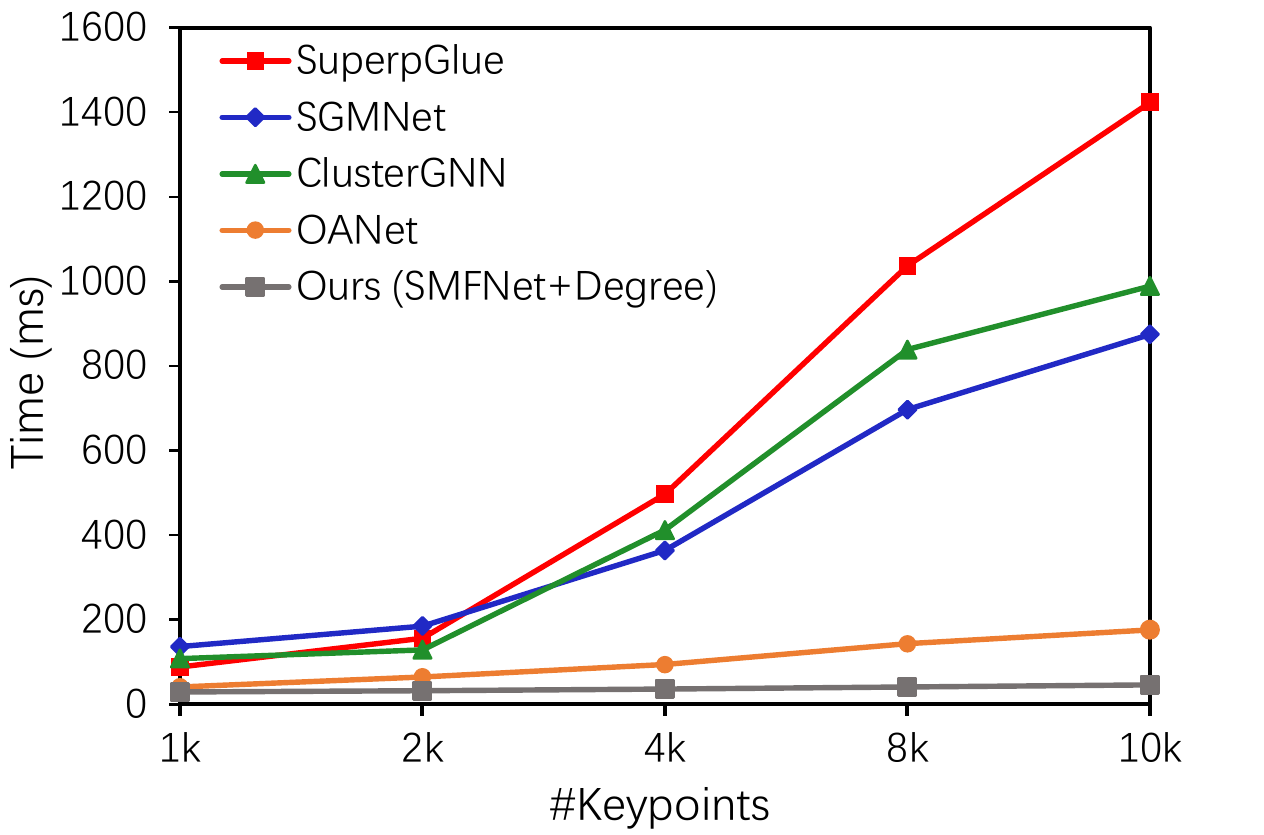}}
    \subfigure[Memory consumption]{
    \includegraphics[width=0.43\textwidth]{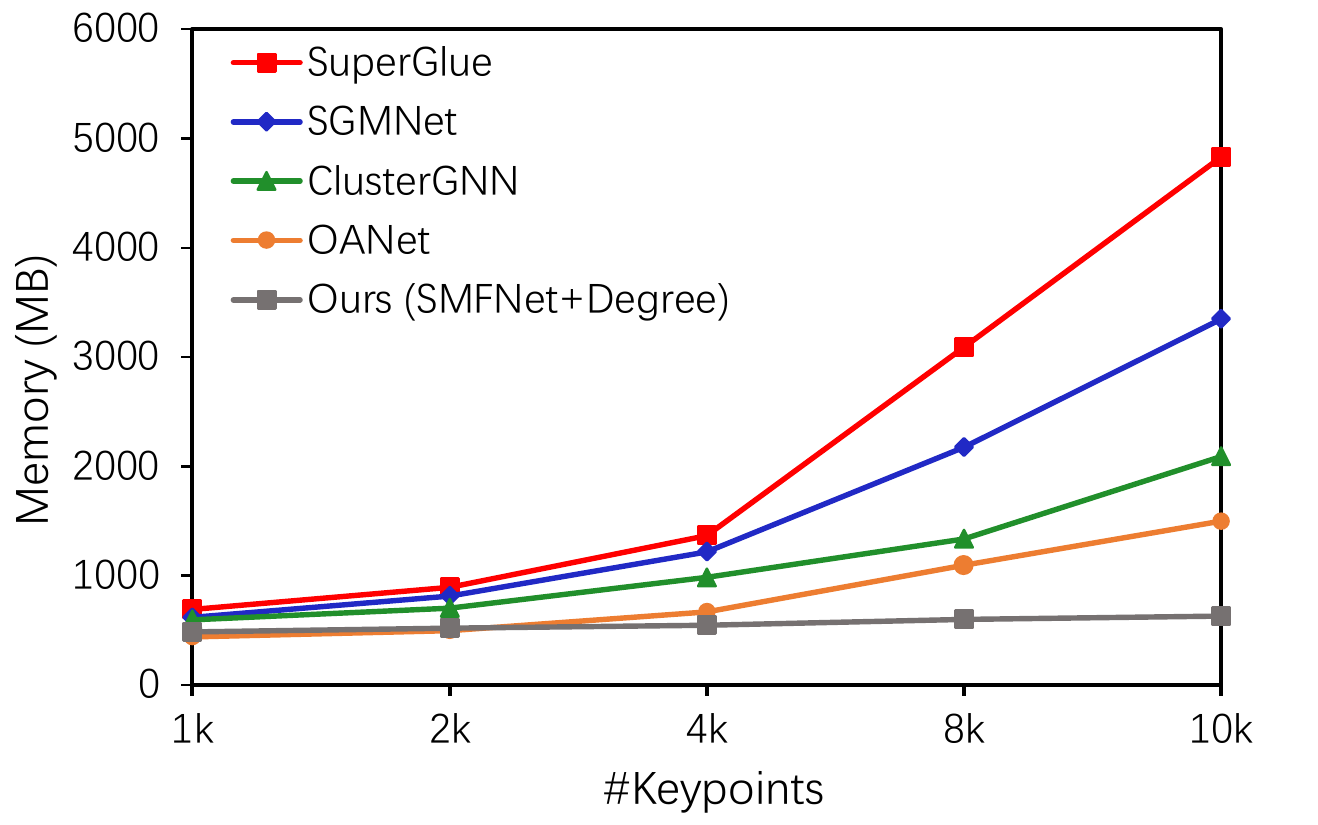}}
    \caption{Efficiency comparison. We report the (a) time and (b) memory consumption with increased number of input keypoints.}
\label{pic:efficiency}
\end{figure}

\subsection{Efficiency}

Improving the matching performance while maintaining efficiency is the main motivation for our approach. Recent 
SGMNet and ClusterGNN both build on SuperGlue to further improve computational efficiency and matching performance. In particular, SGMNet establishes a small set of nodes to reduce the cost of attention. ClusterGNN uses K-means to construct local sub-graphs to save memory cost and computational overhead. To evaluate the 
efficiency of different methods, we test the running time and memory 
consumption on an NVIDIA GTX $3090$ GPU and an Intel(R) Xeon(R) Gold $6226$R CPU @$2.90$GHz with an 
increased number of keypoints. The results are shown in Fig.~\ref{pic:efficiency}.

As shown in Fig.~\ref{pic:efficiency}(a), the runtime of comparing matching methods is relevant to the input number of keypoints, while our approach (SMFNet+\ourmethod) is almost independent of the input number of keypoints (on average $38.9$ ms). In particular, for $10$ K keypoints, the runtime of our method could be negligible compared to SuperGlue. We ascribe such high efficient 
to three reasons: i) the pre-processing of SMFNet 
constrains the number of keypoints; ii) the coarse-to-fine network structure 
discards the stacking of transformer modules, large matrix multiplication, and clustering iterations; iii) \ourmethod can be parallelized with multiple threads, resulting in an average of $7$ ms processing time for an image pair.
Fig.~\ref{pic:efficiency}(b) reports the memory occupation averaged by batch size. Although OANet is more memory efficient with a smaller number of input keypoints, our approach has better time efficiency. For dense detection, our approach only increases 
$12.5\%$ memory usage required by SuperGlue. In addition, while SGMNet and ClusterGNN have made progress in reducing the computation and memory overheads for many keypoints, their 
cost is still relatively high compared to our method.


\begin{table*}
    \caption{The baseline combination of different methods.}
    \label{tab:baseline methods}
    \centering
    \begin{tabular}{@{}ll@{}}
\toprule
    Methods &Baseline \\
\midrule
LoFTR$^{\dag}$ &LoFTR~\citep{sun2021loftr} + DEGENSAC~\citep{chum2005two} \\
OANet$^{\dag}$ &DoG~\citep{lowe2004distinctive} + HardNet~\citep{mishchuk2017working} + OANet~\citep{zhang2019learning} + GuidedMatching \\
KORNIA$^{\dag}$ &DoG + AffNet~\citep{mishkin2018repeatability} + HardNet+KORNIA~\citep{eriba2019kornia} + DEGENSAC \\
Bulit-in matcher$^{\dag}$ &SIFT~\citep{lowe2004distinctive} + Bulit-in matcher~\citep{jin2020image} + DEGENSAC \\

COTR$^{\dag}$ &COTR~\citep{jiang2021cotr} + DEGENSAC \\
DISK$^{\dag}$(depth) &DISK~\citep{tyszkiewicz2020disk} + DEGENSAC \\
AdaLAM$^{\dag}$ &HardNet8~\citep{pultar2020improving} + AffNet~\citep{mishkin2018repeatability} + AdaLAM~\citep{cavalli2020handcrafted} + DEGENSAC \\
SuperGlue$^{\dag}$ &SuperPoint~\citep{detone2018superpoint} + SuperGlue~\citep{sarlin2020superglue} + DEGENSAC\\
Ours &\ourmethod + SuperGlue$^{\dag}$ \\
    \bottomrule
    \end{tabular}
\end{table*}

\begin{table*}[!t]
\centering
\addtolength{\tabcolsep}{2pt}
\renewcommand{\arraystretch}{1.2}
\caption{Stereo and multiview performance on Phototourism dataset.}
\begin{tabular}{@{}lccccc@{}}
\toprule 
\multirow{3}{*}{Method}  &  \multicolumn{2}{c}{Stereo} & \multicolumn{3}{c}{Multiview} \\ \cmidrule(lr){2-3} \cmidrule(lr){4-6}
~ & Num.Inl.$\uparrow$ & mAA(@$10^{\circ}$)$\uparrow$ & Reg.Rat.($\%$)$\uparrow$ & ATE$\downarrow$  & mAA(@$10^{\circ}$)$\uparrow$  
                                        \\ \midrule
LoFTR$^{\dag}$(8K)  & 185.0      & \textbf{0.612} & 94.45   & 0.458  & 0.629  \\
OANet$^{\dag}$(8k) & \textbf{765.3}  & {\uline{0.603}} & 98.92 & \textbf{0.355} & \textbf{0.786} \\
KORNIA$^{\dag}$(8k) &320.5   &0.557 &98.59 &0.409 &0.727 \\\cmidrule(l){1-6}
COTR$^{\dag}$(2k)     & {\uline{522.3}}  & 0.552 & \textbf{99.05} & 0.381 & 0.763  \\
DISK$^{\dag}$(depth)(2k) & 404.1  & 0.512  & 97.94   & 0.412 & 0.730  \\
AdaLAM$^{\dag}$(2k)   & 253.5   & 0.437   & 97.58   & 0.413  & 0.713  \\
SuperGlue$^{\dag}$(2k) & 412.7       & 0.538          & 98.51    & 0.387 & 0.750  \\
\textbf{Ours}(2k) &439.2 &0.567 & {\uline{98.83}} &{\uline{0.375}} &{\uline{0.770}} \\
\bottomrule
\end{tabular}
\label{tab:table-phototourism}
\end{table*}

\begin{table*}[!t] 
\centering
\addtolength{\tabcolsep}{2pt}
\renewcommand{\arraystretch}{1.2}
\caption{Stereo and multiview performance in the PragueParks dataset.}
\begin{tabular}{@{}lccccc@{}}
\toprule 
\multirow{3}{*}{Method} &  \multicolumn{2}{c}{Stereo} & \multicolumn{3}{c}{Multiview} \\ \cmidrule(lr){2-3} \cmidrule(lr){4-6}
~ & Num.Inl.$\uparrow$ & mAA(@$10^{\circ}$)$\uparrow$ & Reg.Rat.($\%$)$\uparrow$ & ATE$\downarrow$  & mAA(@$10^{\circ}$)$\uparrow$  
                                      \\ \midrule
LoFTR$^{\dag}$(8K)  & {\uline{321.65}} & \textbf{0.754} & 74.31  & 6.862  & 0.457  \\
Bulit-in matcher$^{\dag}$(8k) &\textbf{444.0} & 0.537 & 73.95   & {\uline{6.522}}  & 0.481\\
KORNIA$^{\dag}$(8k) & 226.0  & 0.598   & 74.59  &\textbf{6.530}  & 0.470\\\cmidrule(l){1-6}
COTR$^{\dag}$(2k)      & 317.7    & 0.589  & 74.69  & 6.911  & 0.472  \\
DISK$^{\dag}$(depth)(2k) & 246.1   & 0.459  & 74.61   & 6.707  & 0.436 \\
AdaLAM$^{\dag}$(2k)    & 142.1   & 0.636 & {\uline{75.82}}  & 6.84  & 0.468  \\
SuperGlue$^{\dag}$(2k) & 281.5 & 0.726  & \textbf{75.92} & 6.921 & {\uline{0.489}} \\
\textbf{Ours}(2k)  & 275.0 & {\uline{0.744}} & 74.80  & 6.957 & \textbf{0.497} \\
\bottomrule
\end{tabular}
\label{tab:table-pragueparks}
\end{table*}

\begin{table*}[!t]
\centering
\addtolength{\tabcolsep}{2pt}
\renewcommand{\arraystretch}{1.2}
\caption{Stereo and multiview performance in the GoogleUrban dataset.}
\begin{tabular}{@{}lccccc@{}}
\toprule 
\multirow{3}{*}{Method}  &  \multicolumn{2}{c}{Stereo} & \multicolumn{3}{c}{Multiview} \\ \cmidrule(lr){2-3} \cmidrule(lr){4-6}
~  & Num.Inl.$\uparrow$ & mAA(@$10^{\circ}$)$\uparrow$ & Reg.Rat.($\%$)$\uparrow$ & Num.Obs.$\uparrow$  & mAA(@$10^{\circ}$)$\uparrow$  
                                      \\ \midrule
LoFTR$^{\dag}$(8K) & \textbf{722.3} & {\uline{0.405}} & {\uline{88.48}}  & 3.357 & 0.289   \\ 
Bulit-in matcher$^{\dag}$(8k)     & 120.2  & 0.269  & 61.08   & 2.878 & 0.090   \\
KORNIA$^{\dag}$(8k) & 100.0  & 0.301  & 74.65   & 2.849  & 0.164  \\\cmidrule(l){1-6}
COTR$^{\dag}$(2k) & 239.8  & 0.336  & 82.87 & 3.556 & 0.231   \\
DISK$^{\dag}$(depth)(2k)  & 159.7 & 0.276 & 70.09 & 3.361   & 0.127   \\
AdaLAM$^{\dag}$(2k)     & 85.0  & 0.325  & 70.70 & 3.063  & 0.160   \\
SuperGlue$^{\dag}$(2k)  & 260.0  & 0.394 & 87.89   & {\uline{3.605}} & {\uline{0.324}}    \\
\textbf{Ours}(2k)  & {\uline{349.7}} & \textbf{0.424} & \textbf{90.19} & \textbf{3.708} & \textbf{0.344} \\

\bottomrule
\end{tabular}
\label{tab:table-GoogleUrban}
\end{table*}

\subsection{CVPR 2021 Image Matching Challenge}
 Here we further report the performance of \ourmethod on the recent Image Matching Challenge at CVPR 2021.\footnote{https://www.cs.ubc.ca/research/image-matching-challenge/2021}
The challenge features three datasets with two tracks: stereo and multiview.

\vspace{5pt} \noindent\textbf{Comparative Methods.}
The comparison methods can be divided into the 8k keypoint category and the 2k keypoint category (as shown in Table~\ref{tab:baseline methods}). The 8k-keypoint category includes:

\begin{itemize}
    \item LoFTR~\citep{sun2021loftr}: a transformer-based dense matching approach with self and cross attention;
    \item OANet~\citep{zhang2019learning}: a consistency filtering approach that infers the probabilities of correspondences where DoG~\citep{lowe2004distinctive} and HardNet~\citep{mishchuk2017working} are used as detector and descriptor, respectively;
    \item Bulit-in matcher~\citep{jin2020image}: a classic matching approach with SIFT~\citep{lowe2004distinctive} descriptor, Ratio Test, Mutual NN~\citep{lowe2004distinctive} and MEGSAC~\citep{barath2019magsac};
    \item KORNIA~\citep{eriba2019kornia}: an open source computer vision library designed for feature matching, which uses DoG detector and HardNet descriptor with AffNet~\citep{mishkin2018repeatability}.
\end{itemize}
The 2k-keypoint category contains:
\begin{itemize}
    \item COTR~\citep{jiang2021cotr}: a dense matching approach that finds correspondences using points as queries in one of the image pair;
    \item DISK (depth)~\citep{tyszkiewicz2020disk}: an end-to-end local feature framework that searches for many correct feature matches;
    \item AdaLAM~\citep{cavalli2020handcrafted}: an outlier filtering approach with a hierarchical pipeline for outlier detection, which uses the HardNet8~\citep{DBLP} descriptor with AffNet;
    \item SuperGlue~\citep{sarlin2020superglue}: an end-to-end sparse feature matching approach with attention-based context aggregation, which uses SuperPoint~\citep{detone2018superpoint} as the descriptor;
    \item \textbf{Ours}: \ourmethod is applied based on the SuperPoint descriptor and SuperGlue method, which encodes keypoint coordinates for SuperGlue.
\end{itemize}

Except for the bulitin matcher and OANet, the final results of these methods are pre-filtered by DEGENSAC~\citep{chum2005two}. All models are pre-trained in the MegaDepth dataset~\citep{li2018megadepth}. Following the organizer's requirements, we remove overlapping scenes from the dataset. Public datasets~\citep{jin2020image} include ``Phototourism'', ``PragueParks'', and ``GoogleUrban''. Here, we highlight some key results for each dataset.

\vspace{5pt} \noindent\textbf{The Phototourism Dataset.} 
Images of this dataset are collected from multiple sensors captured at different times, from different viewpoints and with occlusions. As shown in Table~\ref{tab:table-phototourism}, \ourmethod outperforms all methods in the category of 2k keypoints on the stereo task and is on par with the second-best method among all entries on the multiview task leaderboard, including those that use 8k keypoints. In particular, compared to the original SuperGlue baseline, \ourmethod improves all evaluation metrics on both tasks.

\vspace{5pt} \noindent\textbf{The PragueParks Dataset.} 
This dataset includes complex natural environmental scenarios. The results are reported in Table~\ref{tab:table-pragueparks}. \ourmethod outperforms all methods in the 2k-keypoint category on both tasks and is on par with the best method among all entries in the multiview task leaderboard.

\vspace{5pt} \noindent\textbf{The GoogleUrban Dataset.} 
This dataset focuses on low overlap ratios between image pairs from street architectures. As shown in Table~\ref{tab:table-GoogleUrban}, \ourmethod outperforms almost all competitors in both tasks, especially with an average of $2\%$ higher than the second best entry in mAA (@$10^\circ$).

\section{Conclusion}
This paper demonstrates the power of geometric-invariant coordinates for learned correspondences. We propose a coordinate encoding approach \ourmethod, which encodes barycentric coordinates based on dynamic seed correspondences (basis points). We show how to encode the coordinates using seed correspondences. \ourmethod can be used as a plug-in applicable to existing descriptors and correspondence networks to improve the matching performance. We also introduce an SMFNet network that can generate reliable seed correspondences. We evaluated the generalizability of \ourmethod in various datasets and achieved competitive results compared to the latest learning-based methods in the Image Matching Challenge 2021. We believe that \ourmethod opens a new door to considering robust feature correspondences from the perspective of coordinate representations. \ourmethod currently cannot handle some extreme cases. To further improve the confidences of seed correspondences under significant occlusions, we plan to use dense matching with multiscale feature fusion instead of sparse coarse matching. For the coordinate deviations caused by a large-scale depth gap, we plan to apply probabilistic selection for encoded coordinates.

\begin{center}
    {\bf \large Appendix}
\end{center}

\noindent{\bf Proof of Theorem 1} Rotation Invariance. \begin{proof}
With an angle $\theta$ and origin, a rotation matrix ${\bm R}$ can be obtained by rotating the points counterclockwise.
\begin{equation}
        {\bm R} = \begin{bmatrix}\cos\theta & -\sin\theta \\ \sin\theta & \cos\theta \end{bmatrix} \,,
\end{equation}
such that
\begin{equation}
    \begin{bmatrix}x_i^{1} \\ y_i^{1}  \end{bmatrix} = {\bm R}\begin{bmatrix}x_i^{0} \\ y_i^{0}  \end{bmatrix}\,.
\end{equation}
Then, for a coordinate axis $\overrightarrow{{\bm x}_{0}^1{\bm x}_{1}^1}$, we have the following.
\begin{equation} \label{eq:7}
    \begin{split}
        \overrightarrow{{\bm x}_{0}^1{\bm x}_{1}^1} &= \begin{bmatrix}x_1^1\\ y_1^1\end{bmatrix} - \begin{bmatrix}x_0^1\\ y_0^1\end{bmatrix} 
        \\
        &=  {\bm R}\left(\begin{bmatrix}x_1^0\\ y_1^0\end{bmatrix} - \begin{bmatrix}x_0^0\\ y_0^0\end{bmatrix}\right)
        \\
        &= {\bm R}\overrightarrow{{\bm x}_{0}^0{\bm x}_{1}^0}
    \end{split}\,.
\end{equation}
Similarly, for another coordinate axis $\overrightarrow{{\bm x}_{0}^1{\bm x}_{2}^1}$, we can derive $\overrightarrow{{\bm x}_{0}^1{\bm x}_{2}^1} = {\bm R} \overrightarrow{{\bm x}_{0}^0{\bm x}_{2}^0}$\,. Then, according to Eq.~(7) and Eq.~(9) in the main text, the transition matrix
\begin{equation} \label{eq:s9}
    \begin{split}
        {\bm T}_1&=
        \begin{bmatrix}
        \Delta{x}_1^1 & \Delta{x}_2^1 \\
        \Delta{y}_1^1 & \Delta{y}_2^1
        \end{bmatrix}\\
        &= \begin{bmatrix}\overrightarrow{{\bm x}_{0}^1{\bm x}_{1}^1},\, \overrightarrow{{\bm x}_{0}^1{\bm x}_{2}^1}\end{bmatrix}\\
        &= {\bm R}\begin{bmatrix}\overrightarrow{{\bm x}_{0}^0{\bm x}_{1}^0},\, \overrightarrow{{\bm x}_{0}^0{\bm x}_{2}^0}\end{bmatrix} \\
        &= {\bm R}{\bm T}_0
    \end{split}\,.
\end{equation} 
Therefore, 
$\forall \bm p_k^1 \in \mathcal{P}_1$\,, its projection ${{\bm p}_k^1}'$ takes the form
\begin{equation} \label{eq:rotation}
  \begin{split}
      {{\bm p}_k^1}' &={\bm T}_1^{-1} \begin{bmatrix}\Delta{x}_k^1\\ \Delta{y}_k^1\end{bmatrix}
      \\
      &= {\bm T}_1^{-1}\begin{bmatrix} x_k^1 - x_0^1\\ y_k^1 - y_0^1\end{bmatrix}
      \\
      &= \left({\bm R}{\bm T}_0\right)^{-1} \left(\bm{R}\begin{bmatrix} x_k^0 - x_0^0\\ y_k^0 - y_0^0\end{bmatrix}\right)
      \\
      &= {\bm T}_0^{-1} {\bm R}^{-1} {\bm R} \begin{bmatrix}\Delta{x}_k^0\\ \Delta{y}_k^0\end{bmatrix}
      \\
      &= {\bm T}_0^{-1} \begin{bmatrix}\Delta{x}_k^0\\ \Delta{y}_k^0\end{bmatrix}
      \\
      &= {{\bm p}_k^0}' 
      \\
  \end{split} \,.
\end{equation}
The proof is complete.
\end{proof}

\noindent{\bf Proof of Theorem 2} Affine Invariance. \begin{proof}
Given an affine matrix $\bm A_{2\times3}$ (hereafter referred to as $\bm A$) and the augmented coordinate vector $\hat{{\bm x}}_i^0=\begin{bmatrix}{\bm x}_i^0 \\ 1\end{bmatrix}$, the affine transformation can be written as ${\bm x}_i^1 = {\bm A}\hat{{\bm x}}_i^0$, \ie, 
\begin{equation}\label{eq:affine_transform}
        \begin{bmatrix}x_i^1 \\y_i^1 \end{bmatrix} =\begin{bmatrix}a_1 & a_2 & t_x \\a_3 & a_4 & t_y \end{bmatrix}\begin{bmatrix}x_i^0 \\y_i^0 \\ 1 \end{bmatrix}
         \,,
\end{equation}
where $\begin{bmatrix}t_x, t_y \end{bmatrix}^{T}$ is a translation vector. According to the translation invariance property, we can simplify Eq.~\eqref{eq:affine_transform} to ${\bm x}_i^1  = {\hat{\bm A}}_{2 \times 2}{\bm x}_i^0$, \ie,
\begin{equation}
        \begin{bmatrix}x_i^1 \\y_i^1 \end{bmatrix} =\begin{bmatrix}a_1 & a_2 \\a_3 & a_4 \end{bmatrix}\begin{bmatrix}x_i^0 \\y_i^0 \end{bmatrix}\,,
\end{equation}
where $\hat{\bm A}_{2\times 2}=\begin{bmatrix}a_1 & a_2 \\a_3 & a_4 \end{bmatrix}$ (hereafter referred to as $\hat{\bm A}$).
Following Eq.~\eqref{eq:7}, we have $\overrightarrow{{\bm x}_{0}^1{\bm x}_{1}^1} = {\hat{\bm A}} \overrightarrow{{\bm x}_{0}^0{\bm x}_{1}^0}$\, and $\overrightarrow{{\bm x}_{0}^1{\bm x}_{2}^1} = {\hat{\bm A}} \overrightarrow{{\bm x}_{0}^0{\bm x}_{2}^0}$\,. Thus, it is easy to derive ${\bm T}_1 = \hat{\bm A}{\bm T}_0$ according to Eq.~\eqref{eq:s9}. Similarly to Eq.~\eqref{eq:rotation}, $\forall \bm p_k^1 \in \mathcal{P}_1$, its projection ${{\bm p}_k^1}'$ takes the form
\begin{equation}
  \begin{split}
      {{\bm p}_k^1}' &={\bm T}_1^{-1} \begin{bmatrix}\Delta{x}_k^1\\ \Delta{y}_k^1\end{bmatrix}
      \\
      &= \left({\hat{\bm A}}{\bm T}_0\right)^{-1} \left(\hat{\bm A}\begin{bmatrix} x_k^0 - x_0^0\\ y_k^0 - y_0^0\end{bmatrix}\right)
      \\
      &= {\bm T}_0^{-1} {\hat{\bm A}}^{-1} {\hat{\bm A}} \begin{bmatrix}\Delta{x}_k^0\\ \Delta{y}_k^0\end{bmatrix}
      \\
      &= {{\bm p}_k^0}'
      \\
  \end{split} \,.
\end{equation}
The proof is complete.
\end{proof}

\bibliographystyle{spbasic} 
\bibliography{egbib}


\end{document}